\documentclass{article}
\usepackage[sglblindworkshop, final]{neurips_2026}
\setcitestyle{square,numbers}
\usepackage[utf8]{inputenc}
\usepackage[T1]{fontenc}
\usepackage{amsmath,amssymb,graphicx,booktabs,microtype}
\usepackage{xcolor}
\usepackage{url}
\usepackage[breaklinks,colorlinks,citecolor=teal,linkcolor=blue,urlcolor=blue]{hyperref}
\usepackage[capitalize,noabbrev,nameinlink]{cleveref}
\usepackage{enumitem}
\usepackage{tcolorbox}
\tcbuselibrary{listings,breakable,skins}
\lstdefinestyle{promptstyle}{basicstyle=\ttfamily\footnotesize,breaklines=true,
  columns=fullflexible,keepspaces=true,showstringspaces=false,
  literate={–}{{-}}1 {—}{{---}}3 {≥}{{$\geq$}}1 {≤}{{$\leq$}}1
  {→}{{$\to$}}1 {×}{{$\times$}}1 {≠}{{$\neq$}}1 {°}{{deg}}3
  {≈}{{$\approx$}}1 {α}{{$\alpha$}}1 {β}{{$\beta$}}1 {²}{{\textsuperscript{2}}}1 {³}{{\textsuperscript{3}}}1 {π}{{$\pi$}}1 {∈}{{$\in$}}1 {−}{{$-$}}1 {∠}{{$\angle$}}1 {∥}{{$\parallel$}}1 {∩}{{$\cap$}}1 {⊥}{{$\perp$}}1}
\tcbset{promptbox/.style={enhanced,breakable,listing only,
  colback=gray!3,colframe=gray!55,colbacktitle=black,boxrule=0.4pt,arc=1mm,
  left=5pt,right=5pt,top=4pt,bottom=4pt,listing options={style=promptstyle}}}

\definecolor{RoleQ}{HTML}{2A5BD7}
\definecolor{RoleV}{HTML}{A66A00}
\definecolor{RoleS}{HTML}{0A7A3D}
\newcommand{\domaintag}[1]{%
  \fcolorbox{gray!55}{gray!12}{\scriptsize\strut\textsc{#1}}}

\tcbset{qualstyle/.style={enhanced,boxrule=0.4pt,arc=0.7mm,
  coltitle=black,fonttitle=\bfseries\scriptsize,
  left=4pt,right=4pt,top=2pt,bottom=2pt,
  toptitle=1pt,bottomtitle=1pt,
  before skip=2.5pt,after skip=2.5pt,
  attach boxed title to top left={xshift=3mm,yshift=-2pt},
  boxed title style={size=small,colback=white,boxrule=0.3pt}}}


\newcommand{\affmark}[1][]{\textsuperscript{#1}}
\newcommand{\affaddr}[1]{#1}

\title{Trinity: Self-Evolving Vision-Language Models\\with a Self-Verifier}
\workshoptitle{Agentic AI for Biological Discovery (AgenticLS)}
\author{%
Youngwan Lee\affmark[1,2]~~~Yong-Ju Lee\affmark[1]~~~Sung Ju Hwang\affmark[2,3]\\ 
\vspace{-0.1in}
\\
\affaddr{\affmark[1]Electronics and Telecommunications Research Institute~(ETRI), South Korea}\\
\affaddr{\affmark[2]Korea Advanced Institute of Science and Technology~(KAIST), South Korea}\\
\affaddr{\affmark[3]DeepAuto.ai, South Korea}
}
\begin{document}
\maketitle
\addtocontents{toc}{\protect\setcounter{tocdepth}{-1}}

\begin{abstract}
Self-evolving vision-language models (VLMs), a form of self-improvement in which a model generates its own training data from unlabeled images, are a promising route toward agents that expand their reasoning capability in an unsupervised manner, without relying on ever-larger annotation budgets.
Existing methods pair a Questioner that proposes problems with a Solver that answers them, but reward both roles mainly by agreement among sampled answers.
Agreement is a weak proxy for truth: it cannot tell whether a question is grounded in the image, whether the proposed reference answer is right, or whether a confident majority is wrong in the same way.
We present \textsc{Trinity}, in which one VLM plays three roles, Questioner, Solver, and Verifier, and the Verifier is a self-verifier: an exponential moving average (EMA) of the policy itself, requiring neither labels nor an external judge.
The Verifier screens every generated question for image grounding and answer correctness before it becomes supervision, scores Solver reasoning against the image, and adjudicates disputes between the reference answer and a strong Solver consensus, correcting the reference and penalizing the Questioner when the consensus is right.
Trained on images alone, Trinity improves Qwen3-VL-8B on mathematical visual reasoning and on science benchmarks with biology content, for example $+8.6$ points on the biology split of SciVQR and $+12.8$ on MathVerse, and its reward dynamics behave as a healthy self-play curriculum should.
These results suggest that a self-evolving multimodal agent can strengthen its scientific reasoning from unlabeled scientific images alone, with the model itself serving as the verifier.
\end{abstract}

\section{Introduction}
Agentic AI for biological discovery asks models to do more than predict: to read figures, plan multi-step analyses, and improve from their own experience.
A recurring theme is self-improvement: an agent generates its own tasks, attempts them, and turns the outcome into training signal, so capability grows without proportional growth in expert annotation.
For VLMs this idea has crystallized as \emph{self-evolving} training from unlabeled images: a Questioner proposes an image-conditioned problem, a Solver answers it, and both are optimized with reinforcement learning (RL) on rewards computed from the model's own outputs.
Because scientific images are abundant while expert question--answer (QA) annotations are scarce, such a loop is attractive for life-science agents.

Text-only self-play established the proposer--solver pattern (Appendix~\ref{app:related}): Absolute Zero~\citep{zhao2025absolutezero} proposes and solves code-verifiable tasks with one model, and R-Zero~\citep{huang2025rzero} co-evolves a Challenger and a Solver with consistency rewards.
For VLMs, VisPlay~\citep{he2026visplay} trains a Questioner and a Reasoner with difficulty and diversity rewards; V-Zero~\citep{wang2026vzero} uses majority-vote pseudo-labels and a dual-track question reward; EvoLMM~\citep{thawakar2025evolmm} makes the consistency reward continuous; iReasoner~\citep{sunil2026ireasoner} adds agreement between reasoning trajectories; MM-Zero~\citep{li2026mmzero} adds a Coder that renders images; Agent0-VL~\citep{liu2025agent0vl} verifies tool-integrated solutions with tool feedback; and DPE~\citep{jia2026blind} steers tool-assisted data generation toward diagnosed blind spots.

Across these systems the dominant source of reward is \emph{agreement}: a question is good if Solver samples disagree at a target rate, and a Solver sample is right if it matches the majority.
It fails as a stand-in for truth in two ways.
First, it is blind to \emph{grounding}: a question that references objects or values absent from the image, or that can be answered from memory without the image, produces disagreement that looks like difficulty, so a consensus-rewarded Questioner drifts toward ill-posed questions.
Second, it is blind to a \emph{wrong reference}: when the Questioner's own answer is wrong, correct Solver samples are penalized, and when Solver samples share a systematic error, the majority is confidently wrong.
Neither failure is visible from answer statistics; both require an evaluator that looks at the image.

\begin{figure}[t]
  \centering
  \includegraphics[width=\linewidth]{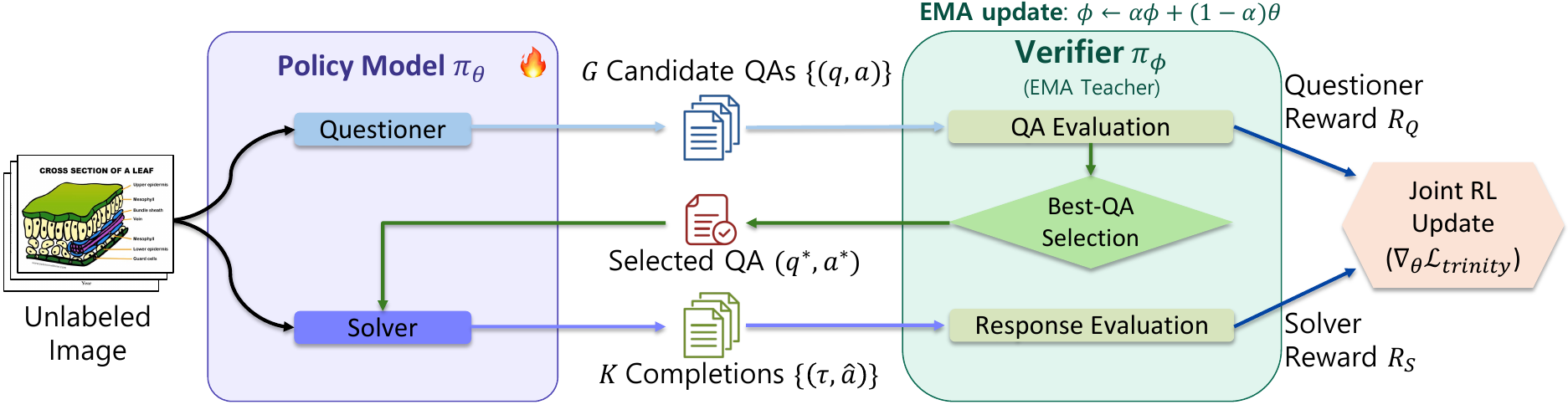}
\caption{\textbf{Trinity training loop.} The policy $\pi_\theta$ as Questioner proposes $G$ candidate QAs $\{(q,a)\}$; the EMA Verifier $\pi_\phi$ screens them and selects $(q^*,a^*)$; the policy as Solver produces $K$ completions $\{(\tau,\hat a)\}$, which the Verifier scores.
$R_Q$ and $R_S$ drive one joint update $\nabla_\theta\mathcal L_{\rm trinity}$, after which $\phi\leftarrow\alpha\phi+(1-\alpha)\theta$.}
  \label{fig:trinity-arc}
\end{figure}

\textsc{Trinity}, named for its three roles in one model, adds such an evaluator without adding labels or an external judge: a self-verifier.
A single VLM plays three roles: the Questioner and Solver share the trainable policy weights and differ only by prompt, while the Verifier is an EMA copy of the same policy that receives no gradients and lags the policy by design.
The Verifier intervenes at three points of the loop (\cref{fig:trinity-arc}): (i) it screens each generated QA for image grounding, answer correctness, and difficulty before it can become Solver supervision; (ii) it scores each Solver reasoning trace against the image; and (iii) when a strong Solver consensus contradicts the reference, it re-solves the problem and, if the consensus wins, corrects the reference and revokes the Questioner's credit.
Our contributions are (1)~a three-role, single-model self-evolving framework whose self-verifier, an EMA copy of the policy, inherits the Solver's gains without a critic, tools, or labels; (2)~verifier-gated rewards that separate genuine difficulty from invalidity and credit reasoning only on correct answers; and (3)~gains on two backbones in mathematical visual reasoning and on scientific benchmarks with biology content.

\section{Method}
\label{sec:method}

\paragraph{Roles and notation.} Let $x$ be an unlabeled image.
We keep two copies of one VLM: the \emph{policy} $\pi_\theta$, which plays the Questioner and the Solver under role-specific prompts and receives gradients, and the \emph{Verifier} $\pi_\phi$, an EMA copy of $\pi_\theta$ that is used only for inference.
The Questioner samples $G$ candidate pairs $(q,a)\sim\pi_\theta(\cdot\mid x)$, where $a$ is a short reference answer.
For the selected pair $(q^*,a^*)$ the Solver samples $K$ completions $(\tau,\hat a)\sim\pi_\theta(\cdot\mid x,q^*)$, where $\tau$ is a chain of thought and $\hat a$ its boxed final answer.
Verifier calls return the signals below as JSON (prompts in Appendix~\ref{app:prompts}).

\paragraph{Verifier QA evaluation and best-QA selection.} Given $(x,q,a)$, the Verifier returns three signals: question factuality $R_{\text{fact-q}}\in\{0,1\}$ (every premise of $q$ is visible in $x$ and $q$ is not answerable without $x$), answer correctness $R_{\text{corr-a}}\in\{0,1\}$, and difficulty $R_{\rm diff}\in\{0.1,0.4,0.7,1.0\}$ from a four-level rubric (observable, single-step, multi-step, conceptual) that caps single-theorem and lookup questions.
Self-contradictory outputs ($R_{\text{fact-q}}=0$, $R_{\text{corr-a}}=1$) are clamped to $R_{\text{corr-a}}=0$.
Among format-valid candidates that pass the gate we select the hardest, $(q^*,a^*)=\arg\max_{g} R^{(g)}_{\rm diff}$ subject to $R^{(g)}_{\text{fact-q}}=R^{(g)}_{\text{corr-a}}=1$, breaking ties at random; if no candidate passes, the image contributes Questioner rollouts only.
The gate is what separates a hard question from a broken one: the Solver fails on both, but only the former becomes supervision.

\paragraph{Response evaluation.} Each Solver completion receives three signals.
Correctness $R_{\rm correct}\in\{0,1\}$ uses a three-tier procedure: (i) a rule-based match against $a^*$ that normalizes LaTeX, units, and ratios; (ii) a numeric hard-mismatch guard that returns $0$ when both answers parse to numbers outside a $1\%$ tolerance, blocking Verifier false positives; and (iii) a text-only Verifier equivalence check for the residual ambiguous cases.
Reasoning quality $R_{\rm reason}\in\{0,0.25,0.5,1\}$ is scored by the Verifier from the image and $\tau$, penalizing visual misreads, invalid logic, and shortcuts.
A rule-based check gives the format indicator $R^{S}_{\rm fmt}\in\{0,1\}$.

\paragraph{Dispute resolution.} If at least $75\%$ of the formatted Solver answers agree on a value that differs from $a^*$, and all of them scored $R_{\rm correct}=0$, the Verifier receives the image, both answers, and one supporting $\tau$, solves the question itself, and picks a side.
If it sides with the Solvers, $a^*$ is overwritten, every $R_{\rm correct}$ is recomputed, and the Questioner's $R_{\text{corr-a}}$ is set to $0$.
Consensus triggers verification; the Verifier, not the vote, decides.

\paragraph{Questioner reward $R_Q$.} We combine a format gate $R^{Q}_{\rm fmt}\in\{0,1\}$ (valid JSON, answer of at most ten words), a factuality gate $R_{\rm fact}=R_{\text{fact-q}}\,(1+R_{\text{corr-a}})/2\in\{0,0.5,1\}$, the Verifier's a-priori difficulty $R_{\rm diff}$, and the Solver-validated challenge $R_{\rm chal}=1-\overline{R_{\rm correct}}$, where the bar averages over the $K$ completions of the selected QA: the empirical failure rate of the current Solver.
Because a question that every Solver answers was not actually hard, difficulty is discounted by challenge, $\widetilde R_{\rm diff}=R_{\rm diff}\,(0.5+0.5\,R_{\rm chal})$, and
\begin{equation}
R_Q=\begin{cases}
-\rho_f, & R^{Q}_{\rm fmt}=0,\\
R_{\rm fact}\bigl(\alpha_Q\,\widetilde R_{\rm diff}+\beta_Q\,R_{\rm chal}\bigr), & \text{otherwise},
\end{cases}
\label{eq:rq}
\end{equation}
with $(\alpha_Q,\beta_Q,\rho_f)=(0.3,0.7,0.5)$.
An ungrounded question ($R_{\text{fact-q}}=0$) earns nothing however much the Solver struggles, a grounded question with a wrong reference earns at most half, and $\beta_Q>\alpha_Q$ makes the empirical failure signal dominant.

\paragraph{Solver reward $R_S$.} Correctness gates reasoning credit, so a fluent but wrong chain of thought earns nothing for its reasoning:
\begin{equation}
R_S=w_c\,R_{\rm correct}+w_r\,R_{\rm reason}\,R_{\rm correct}+w_f\,R^{S}_{\rm fmt},\qquad (w_c,w_r,w_f)=(0.5,0.3,0.2).
\label{eq:rs}
\end{equation}
Truncated or malformed completions stay in the buffer with $R_S=0$, penalizing non-termination.

\paragraph{Joint optimization and EMA.} Within each image and role, rewards are standardized into group-relative advantages $A=(R-\mu)/(\sigma+\varepsilon)$ as in GRPO~\citep{shao2024deepseekmath}.
Questioner and Solver rollouts are pooled and optimized with the sequence-level clipped surrogate of GSPO~\citep{zheng2025gspo}, whose ratio for a completion $o$ is $\rho_o(\theta)=\exp\bigl[\frac{1}{|o|}\sum_t\log\frac{\pi_\theta(o_t\mid o_{<t})}{\pi_{\theta_{\rm old}}(o_t\mid o_{<t})}\bigr]$; $\mathcal L_{\rm trinity}$ averages the clipped loss and an old-policy regularizer over retained rollouts of both roles.
After each outer step the Verifier is refreshed by a Polyak average, $\phi\leftarrow\alpha\phi+(1-\alpha)\theta$ with $\alpha=0.999$.
The Verifier thus inherits the Solver's gains slowly, so the reward landscape moves smoothly and the policy cannot exploit its own quirks as a grader.
At deployment only the Solver policy is used.

\section{Experiments}
\label{sec:experiments}
\paragraph{Setup.} We build Trinity based on Qwen2.5-VL-7B-Instruct~\cite{bai2025qwen25vl} and Qwen3-VL-8B-Instruct~\citep{bai2025qwen3} as backbones, using only the images of Vero-600K~\citep{vero600k}, about 600K images of which one sixth are STEM (science diagrams, geometry, pathology and radiology).
As a data ablation, we also train on OmniScience~\citep{tao2026omniscience}, 1.5M figures from open-access journals and preprints across biology, chemistry, physics, medicine, and other sciences.
Training proceeds in an unsupervised manner: the RL loop consumes raw images only and never reads the corpora's question--answer annotations.
Training details and decoding settings are in Appendix~\ref{app:implementation}; 7B results are in Appendix~\ref{app:additional}.

\paragraph{Benchmarks.} We evaluate on mathematical visual reasoning (MathVision~\citep{wang2024mathvision}, MathVerse~\citep{zhang2024mathverse}, MathVista~\citep{lu2024mathvista}), on MMMU~\citep{yue2024mmmu}, whose college-level questions include biology and medicine, and on three scientific benchmarks chosen for their biological content.
SciVQR~\citep{guo2026scivqr} has a dedicated biology split spanning cell biology, genetics, and ecology; MME-SCI~\citep{ruan2026mmesci} includes biology items; and EMVista~\citep{wen2026emvista} tests electron-microscopy understanding, central to structural and cell biology.

\begin{table}[t]
\centering\footnotesize\setlength{\tabcolsep}{2.6pt}\renewcommand{\arraystretch}{0.9}
\caption{\textbf{Performance comparison with Self-evolving VLMs using Qwen3-VL-8B-Instruct.} Trinity is trained on Vero-600K or OmniScience images.}
\label{tab:main}
\begin{tabular}{lccccccc}
\toprule
Model & MathVision & MathVerse & MathVista & MMMU & EMVista & SciVQR-Bio & MME-SCI \\
\midrule
\textcolor{gray}{Qwen3-VL-8B-Instruct} & \textcolor{gray}{52.3} & \textcolor{gray}{58.2} & \textcolor{gray}{76.7} & \textcolor{gray}{57.0} & \textcolor{gray}{44.8} & \textcolor{gray}{56.7} & \textcolor{gray}{19.0} \\
MM-Zero~\citep{li2026mmzero} & 39.6 & 45.1 & 67.2 & 58.3 & 45.0 & 62.5 & 17.6 \\
DPE~\citep{jia2026blind} & 53.9 & 57.2 & 76.2 & 69.1 & 44.1 & 60.8 & 20.4 \\
\textbf{Trinity} (Vero-600K) & 57.8 & \textbf{71.0} & \textbf{79.6} & 69.6 & \textbf{47.0} & \textbf{65.3} & 20.3 \\
\textbf{Trinity} (OmniScience) & \textbf{58.2} & 70.6 & \textbf{79.6} & \textbf{70.7} & \textbf{47.0} & 62.5 & \textbf{21.9} \\
\bottomrule
\end{tabular}
\end{table}
\paragraph{Main results.} Trained on Vero-600K images alone, Trinity improves the base model on every benchmark (\cref{tab:main}): MathVerse $+12.8$, MMMU $+12.6$, SciVQR-Bio $+8.6$, EMVista $+2.3$, MME-SCI $+1.3$.
Both released self-improving models on this backbone fall short.
MM-Zero, trained on rendered synthetic images, falls below the base on all three math benchmarks and barely moves the science ones (EMVista $+0.3$), so a synthetic curriculum does not transfer.
DPE, which targets weaknesses with external tools and diagnostic benchmarks, matches Trinity on MMMU but trails it by $13.8$ points on MathVerse and does not improve EMVista ($-0.7$).
Trinity's margin is largest where a verified, image-grounded curriculum should matter: multi-step diagram reasoning and micrograph reading.
The OmniScience variant, trained on scientific figures alone, also improves every benchmark and is statistically indistinguishable from the Vero-600K variant on science benchmarks.

\begin{figure}[t]
  \centering
  \includegraphics[width=0.9\linewidth]{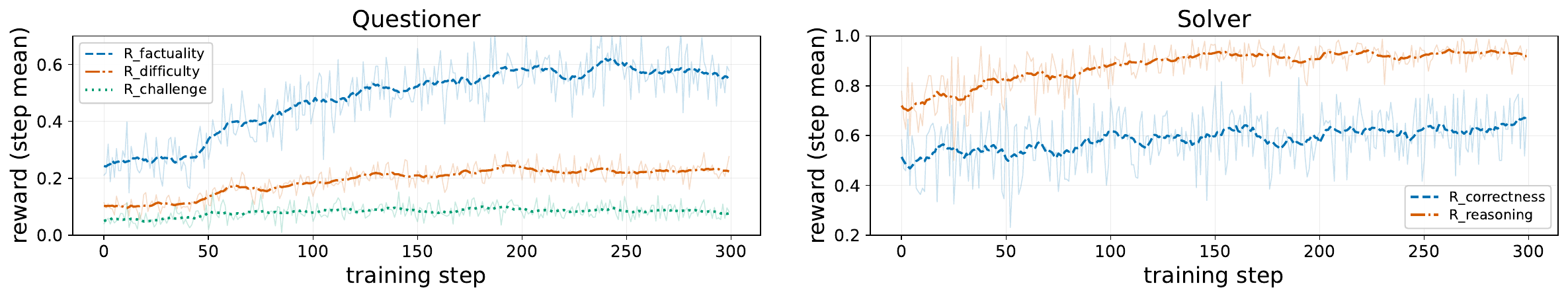}
\caption{\textbf{Reward dynamics for Qwen3-VL-8B on OmniScience~\cite{tao2026omniscience} (steps 0--299).}
}
  \label{fig:reward-trajectory}
\end{figure}

\paragraph{Training dynamics.} \cref{fig:reward-trajectory} plots the per-step mean of each reward component.
Averaged over the first and last 50 steps, question factuality $R_{\rm fact}$ increases from $0.27$ to $0.57$ and Verifier-rated difficulty $R_{\rm diff}$ from $0.11$ to $0.23$, while Solver correctness $R_{\rm correct}$ rises from $0.53$ to $0.63$ and reasoning quality $R_{\rm reason}$ from $0.76$ to $0.93$.
The challenge reward $R_{\rm chal}$ remains below $0.09$, for two reasons.
First, $R_{\rm chal}$ is assigned only to the selected candidate per image, so the plotted mean equals $p_{\rm sel}\,\mathbb E[R_{\rm chal}\mid \text{selected}]$, where $p_{\rm sel}$ is the selection rate; with $p_{\rm sel}$ rising from $0.16$ to $0.23$ and a selected-only mean of $0.36$ in both windows, the products $0.057$ and $0.083$ match the plot.
Second, $R_{\rm chal}$ is a non-stationary target: every question that earns it trains the Solver, which lowers the failure rate on similar questions.
A flat $R_{\rm chal}$ with rising $R_{\rm fact}$ and $R_{\rm diff}$ is thus the expected signature of a curriculum that stays just ahead of the Solver.
Qualitative examples are in Appendix~\ref{app:qualitative}.

\section{Conclusion}
\label{sec:conclusion}
We presented Trinity, a self-evolving framework in which a single VLM acts as Questioner, Solver, and Verifier, with the Verifier instantiated as an EMA replica of the policy that requires neither labels nor separate training.
The Verifier screens generated questions, scores reasoning against the image, and adjudicates disputes, turning self-generated supervision into a curriculum that rewards genuine difficulty.
Trained on unlabeled images alone, Trinity improves two backbones and outperforms released self-improving models on the same backbone.
However, the Verifier is the backbone itself, so its judgements are only as reliable as the backbone's perception and reasoning, and keeping three roles co-resident is memory-intensive.
Future work includes stronger backbones as drop-in Verifiers and alternating optimization of the two trained roles; Appendix~\ref{app:limitations} discusses further limitations.

\vspace{1pt}\noindent{\scriptsize\textbf{Acknowledgments.} This work was supported by Institute of Information \& communications Technology Planning \& Evaluation (IITP) grants funded by the Korea government (MSIT) (No.~2022-0-00871, Development of AI Autonomy and Knowledge Enhancement for AI Agent Collaboration) and Electronics and Telecommunications Research Institute (ETRI) grant funded by the Korean government (26CB1200, Development and Application of Science-Specialized Multimodal Foundation Models).\par}

\clearpage
\bibliographystyle{ieeenat_fullname}
\bibliography{main}
\clearpage
\appendix
\addtocontents{toc}{\protect\setcounter{tocdepth}{2}}
\renewcommand{\contentsname}{Appendix Contents}
\tableofcontents
\clearpage

\section{Implementation details}
\label{app:implementation}
\subsection{Training data and supervision}
Vero-600K comprises approximately 600,000 examples from 59 source datasets, balanced across six visual-task categories of about 101,000 rows each: captioning and instruction following, chart and OCR, counting and grounding, knowledge and recognition, spatial reasoning and action, and STEM~\citep{vero600k}.
We chose it because, beyond general visual reasoning, it carries a substantial scientific component.
The STEM category (16.7\% of the corpus) contains science diagrams and textbook questions (AI2D, TQA, MMK12), geometry and mathematics (Geo170K, GeomVerse, GeoQA+, CoSyn-math, WeMath), abstract reasoning (RAVEN), and medical imaging from pathology and radiology (PathVQA, VQA-RAD); the chart category adds figures from arXiv papers (ArXivQA), scientific chart QA (ReachQA, ECD), and infographics.
Trinity's RL data loader reads only the images; it does not use the associated questions, reference answers, or reasoning traces.
The corpus is science-inclusive rather than biology-specialized, and a fixed training-step budget samples this image pool without covering a full epoch over all 600,000 images.

\paragraph{OmniScience (data ablation).} To test whether a science-only corpus changes the outcome, the 8B model is also trained on the images of OmniScience~\citep{tao2026omniscience}, a corpus of about 1.5M figures cropped from open-access papers (25 sources including high-impact journals and arXiv, bioRxiv, and medRxiv) spanning more than 300 sub-disciplines in biology, chemistry, physics, materials science, medicine, ecology, pharmacy, mathematics, and computer science; the 371 training shards contain 1,520,179 rows.
Captions, contexts, and sub-figure annotations are ignored; only the figure images are used.
The training script is identical to the Vero-600K one except for the data path (same warm-up checkpoint, steps, batch, $K$, LoRA, and EMA settings), so the two rows of \cref{tab:main} differ only in the image distribution.
OmniScience figures are larger and elicit longer generations, which is why the 8B configuration uses two images per worker.

\paragraph{Qwen3-VL-8B initialization.} The 8B model starts from a format-warmed Qwen3-VL-8B-Instruct checkpoint.
The warm-up contains 300 examples sampled from Vero-600K: original dataset questions and reference answers are combined with reasoning generated by Qwen3-VL-8B-Thinking.
The generator caps teacher completions at 4,096 tokens and retained reasoning at 2,500 characters, with temperature $0.7$, top-$p=0.95$, and sampling seed 42.
It constructs a Solver-format response with the teacher reasoning and the dataset reference in a boxed answer.
This is supervised initialization, even though its purpose is learning the thinking/answer format, and it must be accounted for in comparisons with an unmodified Instruct checkpoint.

Warm-up uses three epochs, LoRA rank 32 and scaling parameter 64, dropout $0.05$, learning rate $2\times10^{-5}$, weight decay $0.1$, cosine decay with warm-up ratio $0.1$, and effective batch size eight (one device, batch size one, eight accumulation steps).
The adapter is merged before RL.
The 7B configuration starts directly from Qwen2.5-VL-7B-Instruct without this additional warm-up.

\subsection{Per-backbone RL settings}
All runs use a single node with eight NVIDIA A100 80\,GB GPUs, one worker per GPU, bfloat16, frozen backbone weights with trainable language-model LoRA adapters, frozen vision encoder and merger, and gradient checkpointing.
LoRA excludes the embedding and output-head modules.
The common learning rate is $10^{-5}$ with weight decay $0.1$, cosine scheduling, and five warm-up steps.
Each outer step performs one optimization epoch over retained rollouts.
Both use $G=4$ Questioner candidates, EMA coefficient $0.999$, GSPO clipping bounds $1-3\times10^{-4}$ and $1+4\times10^{-4}$, and regularization coefficient $0.10$.
The reward coefficients are those of Eqs.~\eqref{eq:rq} and~\eqref{eq:rs}; there is no length reward.

\paragraph{Qwen2.5-VL-7B-Instruct.} The launcher specifies 500 outer steps, four images per worker (32 globally), and $K=6$ Solver samples for each selected QA.
LoRA rank is 32 and scaling parameter is 64, with zero dropout.
Questioner and Solver generation budgets are each 16,384 tokens; the Verifier budget is 8,192.
Image resolution is bounded between $256\times28^2$ and $1024\times28^2$ pixels.
The vLLM context limit is 20,480, memory-utilization setting $0.25$, and generation sub-batch size four.
FlashAttention-2 is enabled.

\paragraph{Qwen3-VL-8B-Instruct.} Both 8B configurations (Vero-600K and OmniScience, the latter analyzed in \cref{fig:reward-trajectory}) use 300 outer steps, two images per worker (16 globally), and $K=4$.
LoRA rank and scaling parameter are both 32, with zero dropout.
Questioner and Solver generation budgets are each 12,288 tokens; the Verifier budget is 8,192.
Image resolution is bounded between $256\times32^2$ and $1024\times32^2$ pixels.
The vLLM context limit is 20,480, memory-utilization setting $0.30$, and generation sub-batch size two.
This configuration uses SDPA rather than FlashAttention-2.

\paragraph{Sampling and memory.} Thinking-mode generation uses temperature $1.0$, top-$p=0.95$, and top-$k=20$ without a presence penalty.
Non-thinking response evaluation uses temperature $0.7$, top-$p=0.8$, and top-$k=20$.
Instruct models use a thinking prefix in their chat template.
The policy and EMA weights are swapped into the rollout engine as needed.
EMA is updated on merged policy weights before the trainable adapter is unmerged; it is not an average of LoRA factors.
The stored EMA copy and its inference calls incur memory and compute costs, but no optimizer state.

\subsection{Reward and optimization details}
Format validation requires the expected thinking and output structure; a Questioner emits JSON with \texttt{question} and \texttt{answer}, and a Solver emits a boxed answer.
The Questioner answer must be short (at most ten words) and pass additional sentence/placeholder checks.
Truncated or malformed Solver outputs are retained as low-reward examples rather than silently discarded.

The correctness pipeline first handles exact and normalized matches, including numeric forms.
A hard guard rejects clean numeric mismatches outside a 1\% relative tolerance (with a small absolute floor) unless they satisfy the supported percentage equivalence.
Ambiguous residual pairs go to the text-only equivalence prompt.
This checks agreement with the current reference, not independent factual truth.
Image-conditioned dispute evaluation supplies a separate opportunity to revise that reference.

A dispute requires at least two formatted Solver answers and a normalized majority of at least $\lceil0.75n\rceil$ among them.
It fires only if the majority differs from the reference and every majority member initially has correctness zero.
Disputes use at least two retries, independently of the ordinary Verifier retry budget of zero.
If no parsable verdict is obtained after the retries, the implementation adopts the majority.
An override also sets the selected Questioner's $R_{\text{corr-a}}$ to zero and recomputes $R_{\rm fact}$, so it halves rather than eliminates that candidate's reward.
Solver correctness is recomputed through the same rule/guard/fallback pipeline.
The challenge calculation uses the per-image correctness list accumulated for evaluable Solver responses; format-skipped responses need not enter that denominator.

For role $r\in\{Q,S\}$, each image defines a group with mean $\mu^r$ and population standard deviation $\sigma^r$.
Advantages use $(R-\mu^r)/(\sigma^r+\varepsilon)$ with $\varepsilon=10^{-8}$.
When $\sigma^r<\varepsilon$, a group with negative mean instead uses its raw rewards as advantages; a group with nonnegative mean gets zero advantages.
The implementation omits rollouts with $|A|\leq10^{-6}$, then averages over retained, non-padding rollouts per worker.
Role influence therefore depends on surviving rollout counts; the loss is not an equally weighted sum of role means.
Distributed workers use zero-loss padding to align their update schedules.

For token log-ratio $\ell_t=\log\pi_\theta(o_t\mid o_{<t})-\log\pi_{\theta_{\rm old}}(o_t\mid o_{<t})$, the implemented per-rollout loss is
\begin{equation}
\mathcal L_o=-\min\{\rho_oA_o,\operatorname{clip}(\rho_o,1-\epsilon_-,1+\epsilon_+)A_o\}
+\frac{0.10}{|o|}\sum_t\bigl(e^{\ell_t}-\ell_t-1\bigr).
\end{equation}
The regularizer is computed against rollout-policy log probabilities, not a separately frozen initial-policy reference.
Gradients are taken only through the current policy.

\subsection{Inference and benchmark setup}
\label{app:inference}
At evaluation, only the merged Solver policy is needed; Questioner generation, Verifier calls, and reference adjudication belong to training.
The Qwen3-VL evaluation wrapper uses a thinking-prefix chat template, temperature $1.0$, top-$p=0.95$, top-$k=20$, presence penalty $1.5$, and a default maximum of 20,480 new tokens.
The Qwen2.5-VL wrapper provides greedy and thinking modes: thinking mode uses temperature $1.0$, top-$p=0.95$, top-$k=20$ and a default maximum of 20,480 new tokens, and inserts a thinking prefix; its task names select the corresponding answer-format prompt.

\paragraph{Benchmarks in \cref{tab:main}.}
\begin{itemize}[leftmargin=1.2em,itemsep=1pt,topsep=2pt]
\item \emph{MathVision}~\citep{wang2024mathvision} contains competition mathematics problems with diagrams across 16 subjects and five difficulty levels; we use the full test split (3,040 items).
\item \emph{MathVerse}~\citep{zhang2024mathverse} re-renders visual math problems into versions that vary how much information is carried by the diagram versus the text, testing whether the model actually reads the figure; we use the testmini split (3,940 items).
\item \emph{MathVista}~\citep{lu2024mathvista} aggregates mathematical reasoning over charts, plots, geometry, and natural images; we use the 1,000-item testmini split.
\item \emph{MMMU}~\citep{yue2024mmmu} is a college-level multi-discipline benchmark of image-based multiple-choice and open questions spanning biology, health and medicine, the physical sciences, engineering, and the humanities; we use the validation (val) set of 900 items.
\item \emph{EMVista}~\citep{wen2026emvista} is a 1,850-item multiple-choice benchmark of electron-microscopy understanding (morphology, density, and spatial organization of imaged structures).
\item \emph{SciVQR}~\citep{guo2026scivqr} is a multidisciplinary scientific-reasoning benchmark with per-discipline splits; we use the biology split (360 items: 303 multiple-choice and 57 open-response, covering cell biology, genetics, and ecology).
\item \emph{MME-SCI}~\citep{ruan2026mmesci} tests multimodal scientific reasoning across mathematics, physics, chemistry, and biology; we use the English version (1,019 items: 570 multiple-choice and 449 open-response).
\end{itemize}

\paragraph{Evaluation protocol.} MathVision, MathVerse, MathVista, and MMMU follow the lmms-eval protocol with our thinking-aware answer extraction: thinking traces are separated from final answers before extraction, and multiple-choice items are scored by rule.
The three science benchmarks are single-run ($K{=}1$) values with a 40,960-token generation budget for every model. the EMVista value of the Vero-600K model is the first run of a seeded four-sample evaluation, which we verified reproduces independent single runs exactly for the OmniScience model.
Open-response items in SciVQR and MME-SCI are scored with the official SciVQR judge prompt served by a local Qwen3.5-27B judge (the official judge is Qwen2.5-72B-Instruct; judge failures are not counted as errors).
Sampling follows each model's public recommendation (Trinity: temperature $1.0$, top-$p$ $0.95$; Qwen3-VL-8B-Instruct and DPE: $0.7$/$0.8$; MM-Zero: $0.6$/$0.95$).
DPE is the released \texttt{DPE\_v3} checkpoint (three diagnostic--training rounds on Qwen3-VL-8B-Instruct), whose generation configuration matches the base model, so it is evaluated under exactly the base model's conditions.

\subsection{Interpreting the plotted statistics}
The trajectory in \cref{fig:reward-trajectory} covers 19,264 Questioner candidates and 16,460 Solver rollouts over steps 0--299 of Qwen3-VL-8B trained on OmniScience; the Vero-600K model behaves the same way with smaller magnitudes ($R_{\rm fact}$ $0.19\to0.45$, $R_{\rm correct}$ $0.64\to0.71$, $R_{\rm chal}$ $0.029\to0.055$).
Curves use per-step means and a 15-step moving average.
First/last-window statistics in the main text average the first/last 50 step means.
Selected-only challenge statistics pool selected candidates within each window.
Since every step contains 64 Questioner candidates (one resumed step contains 128), the all-candidate mean coincides with the pooled candidate mean up to rounding.
The selection fraction times selected-only mean reproduces the plotted challenge average, up to rounding.

The plotted difficulty is the Verifier's $R_{\rm diff}$, not $\widetilde R_{\rm diff}$, and the plotted reasoning score is $R_{\rm reason}$, not $R_{\rm reason}\,R_{\rm correct}$.
Factuality aggregates candidates that never reach successful verification, so improving format compliance can increase its mean.
Reward trends alone cannot separate better reasoning from changing task selection or changing evaluator behavior; independent benchmarks serve that purpose.

\section{Additional experiments}
\label{app:additional}
\begin{table}[h]
\centering\small\setlength{\tabcolsep}{3pt}
\caption{\textbf{Qwen2.5-VL-7B-Instruct results (accuracy, \%)} against published self-evolving methods on the same backbone. Mathematical numbers for VisPlay~\citep{he2026visplay}, iReasoner~\citep{sunil2026ireasoner}, and V-Zero~\citep{wang2026vzero} are taken from the respective papers; all other values are measured by us. Science benchmarks use a single run with the same prompt for all three models and the judge scoring of Appendix~\ref{app:inference}. VisPlay, iReasoner, and V-Zero have not released checkpoints, so we could not evaluate them on the science benchmarks (`--').}
\label{tab:7b}
\begin{tabular}{lcccccc}
\toprule
Model & MathVision & MathVerse & MathVista & EMVista & SciVQR-Bio & MME-SCI \\
\midrule
\textcolor{gray}{Qwen2.5-VL-7B-Instruct}~\citep{bai2025qwen25vl} & \textcolor{gray}{26.6} & \textcolor{gray}{50.0} & \textcolor{gray}{70.7} & \textcolor{gray}{43.9} & \textcolor{gray}{38.3} & \textcolor{gray}{11.1} \\
VisPlay~\citep{he2026visplay} & \textbf{31.2} & 39.1 & -- & -- & -- & -- \\
iReasoner~\citep{sunil2026ireasoner} & 25.3 & 45.9 & 69.7 & -- & -- & -- \\
V-Zero~\citep{wang2026vzero} & 27.0 & 42.6 & 69.2 & -- & -- & -- \\
EvoLMM~\citep{thawakar2025evolmm} & 24.8 & 44.9 & 70.5 & 46.3 & 38.3 & \textbf{12.0} \\
\textbf{Trinity} (Vero-600K) & 27.7 & \textbf{51.5} & \textbf{71.9} & \textbf{46.9} & \textbf{41.1} & 11.7 \\
\bottomrule
\end{tabular}
\end{table}
\paragraph{Qwen2.5-VL-7B-Instruct.} \cref{tab:7b} compares Trinity with four self-evolving methods that report results on the same Qwen2.5-VL-7B-Instruct~\citep{bai2025qwen25vl} backbone.
Trinity improves the base model on all three mathematical benchmarks (MathVision $+1.1$, MathVerse $+1.5$, MathVista $+1.2$) and is the only method in the comparison that does not fall below the base on MathVerse or MathVista: VisPlay~\citep{he2026visplay} reports the highest MathVision score but a drop of $10.9$ points on MathVerse, and iReasoner~\citep{sunil2026ireasoner}, V-Zero~\citep{wang2026vzero}, and EvoLMM~\citep{thawakar2025evolmm} all regress on MathVerse by $5$ to $7$ points.
The gains on 7B are smaller than on Qwen3-VL-8B~\citep{bai2025qwen3}, which is consistent with the Verifier being the backbone itself: a weaker backbone screens and grades less accurately, so the loop has less reliable supervision to work with.
Published baselines were evaluated by their authors under their own protocols, so the comparison is indicative rather than matched.

On the science benchmarks, where we could evaluate the base model, EvoLMM, and Trinity under one protocol, Trinity scores highest on EMVista ($+3.0$ over the base) and SciVQR-Bio ($+2.8$), while EvoLMM is marginally ahead on MME-SCI ($+0.3$ over Trinity).
Two protocol details matter for reading the science columns.
All three models received the same prompt that asks for a boxed final answer; Trinity is decoded with its recommended sampling (temperature $1.0$, top-$p$ $0.95$) whereas the base model and EvoLMM are decoded greedily, following their release settings.

\paragraph{Note on baselines.} With the exception of EvoLMM, none of the self-evolving methods in \cref{tab:7b} releases a reproducible checkpoint at the time of writing, so they cannot be evaluated on the science benchmarks of \cref{tab:main}; for VisPlay, iReasoner, and V-Zero the comparison is therefore restricted to the mathematical benchmarks reported in their papers.
This is also why the main comparison in \cref{tab:main} is carried out on Qwen3-VL-8B-Instruct, for which two self-improving checkpoints (MM-Zero~\citep{li2026mmzero} and DPE~\citep{jia2026blind}) are publicly available and could be re-evaluated under our protocol.

\section{Related work}
\label{app:related}
\paragraph{Self-play for reasoning without labels.} Absolute Zero~\citep{zhao2025absolutezero} lets one model propose and solve program-verifiable tasks, using code execution as ground truth; R-Zero~\citep{huang2025rzero} co-evolves a Challenger and a Solver from zero data and rewards the Challenger for questions on which the Solver is maximally uncertain; Self-Rewarding LMs~\citep{yuan2024selfrewarding} use the policy itself as an LLM-as-a-judge.
Trinity brings the proposer--solver pattern to images, where no executor exists to certify answers, and replaces self-judging by the current policy with judging by a slowly moving EMA copy.

\paragraph{Self-generated visual curricula.} VisPlay~\citep{he2026visplay} trains image-conditioned question generation and reasoning jointly with difficulty and diversity objectives.
V-Zero~\citep{wang2026vzero} contrasts intuitive and reasoned answers to shape questions and uses majority-derived targets for Solver learning.
EvoLMM~\citep{thawakar2025evolmm} makes self-consistency rewards continuous, and iReasoner~\citep{sunil2026ireasoner} adds trajectory-level agreement to outcome feedback.
These are distinct reward designs, but all derive supervision from statistics of the Solver's own answers.
Trinity instead screens the generated QA against the image before Solver rollouts and revises conflicting references afterward, with consequences for both roles' rewards.

\paragraph{Multi-role and tool-grounded self-improvement.} Vision-Zero~\citep{wang2026visionzero} builds competitive visual games and alternates self-play with policy optimization.
MM-Zero~\citep{li2026mmzero} adds a Coder that renders images from generated concepts, integrating execution feedback and visual verification.
Agent0-VL~\citep{liu2025agent0vl} couples tool-integrated solving with structured verification and self-repair.
Active Zero~\citep{he2026activezero} adds a Searcher that retrieves visual inputs adapted to the learner, whereas Trinity uses a fixed image corpus.
DPE~\citep{jia2026blind}, which also starts from Qwen3-VL-8B-Instruct and Qwen2.5-VL-7B-Instruct, closes a diagnosis--generation--training loop in which multiple agents annotate unlabeled data with web search and image-editing tools and the data mixture is re-targeted at diagnosed weaknesses after each round; its supervision therefore comes from external tools and diagnostic benchmarks, whereas Trinity's comes from a Verifier that is a copy of the model itself.
Multi-role learning and explicit verification therefore predate Trinity; its distinction is EMA-based QA screening, image-conditioned response evaluation, and dispute feedback over an existing image corpus, without a rendering or tool-execution environment.
Unlike tool execution or laboratory observations, the EMA Verifier offers no independent source of factual authority, which we regard as the main limitation of the approach.

\paragraph{Learning verification.} ADPO~\citep{qiu2026adpo} jointly optimizes generation and verification within one policy, using preference-verification rewards and separate advantages with gradient masks.
Trinity does not directly optimize a verification objective: its Verifier follows the policy through EMA and evaluates both newly generated QA pairs and Solver responses.
This also means that improved solving need not imply improved verification.

\paragraph{Scientific reasoning and life-science agents.} Innovator-VL~\citep{wen2026emvista} studies scientific multimodal training and introduces EMVista's microscopy evaluation.
SciVQR~\citep{guo2026scivqr} includes annotated scientific reasoning across disciplines, including a biology split, and MME-SCI~\citep{ruan2026mmesci} probes scientific knowledge and multimodal reasoning with explicit biological content.
Trinity addresses the training-data generation, curation, and feedback components of scientific agents.
Validating its use in biological discovery requires more than general benchmark gains: evidence-grounded biological judgments and downstream scientific utility must be measured directly.

\section{Limitations and Future Work}
\label{app:limitations}
Each limitation below is paired with the experiment or extension that would address it.

\paragraph{Verifier reliability.} The Verifier is an EMA copy of the policy, so it has no source of factual authority independent of the model being trained.
Errors that the policy and the Verifier share are invisible to the loop, the rubric-based factuality, difficulty, and reasoning judgements are model outputs rather than certified labels, and dispute resolution falls back to the Solver majority when no parsable verdict is obtained.
In audits of training rollouts we observed both false negatives (grounded questions marked ungrounded) and false positives (lenient reasoning scores).
Because the Verifier is the backbone itself, its accuracy is bounded by the backbone's own perception and reasoning: a 7B or 8B model that misreads a chart as a Solver will misread it as a Verifier too.
Two directions follow.
First, since Trinity requires no Verifier-specific training, every stronger open VLM can be dropped in as a new backbone and should yield a more accurate Verifier at no extra cost; pairing a fixed Solver with Verifiers of different sizes would measure how much of the gain is attributable to Verifier accuracy, and an independently annotated sample of Verifier decisions would calibrate the judgements the reward relies on.
Second, lightweight external anchors, such as retrieval over the paper a figure comes from, a symbolic checker for numeric answers, or, in the laboratory setting envisioned by the workshop, an experimental readout, could be inserted at the dispute-resolution stage without changing the rest of the loop.

\paragraph{Memory footprint and training cost.} Trinity keeps three roles resident during training: the trainable policy $\pi_\theta$ with its optimizer state, the EMA Verifier $\pi_\phi$ as a full second weight copy, and the vLLM rollout engine into which policy and Verifier weights are alternately swapped.
Even with LoRA adapters, a frozen vision encoder, and gradient checkpointing, this footprint is larger than that of a Questioner--Solver loop without a separate Verifier, and it is the reason our experiments required eight 80\,GB GPUs (Appendix~\ref{app:implementation}); sharing one engine by weight swapping avoids a third replica but adds swap latency to every outer step.
Wall-clock time is dominated by generation rather than by the gradient step, since each outer step performs Questioner generation, Verifier QA screening, Solver generation, Verifier response evaluation, and dispute resolution before a single policy update: for the 8B backbone one outer step takes on the order of ten to twenty minutes on eight A100 GPUs, and 300 steps take several days.
Verifier calls scale with $G$ and $K$, so the loop is more expensive per training example than supervised RL on annotated data, which is the price of not needing annotations.
Cheaper Verifier instantiations, for example a quantized EMA copy or a Verifier that shares the vision encoder with the policy, and batching Verifier calls across the $G$ candidates are the natural ways to reduce both costs.

\paragraph{Joint versus alternating optimization.} Trinity updates the Questioner and the Solver in a single joint step over their pooled rollouts, whereas prior self-evolving VLMs such as VisPlay~\citep{he2026visplay} optimize the two roles in separate phases, holding one fixed while the other is trained.
Joint optimization is simpler to run end to end and lets both roles improve from every batch, but it couples their gradients through the shared LoRA weights: a step that helps the Solver can perturb the Questioner and vice versa, and role influence on the update depends on how many rollouts of each role survive the advantage filter.
A controlled comparison between the joint update used here and an alternating schedule that freezes one role while the other trains, under matched compute, would establish whether this coupling helps or hurts the final Solver.

\paragraph{Scale and backbone coverage.} Our experiments cover two backbones in the 7--8B range from a single model family (Qwen2.5-VL-7B-Instruct and Qwen3-VL-8B-Instruct).
Because the Questioner, Solver, and Verifier all depend on the same backbone's prompt-following and reasoning behaviour, the results do not by themselves establish that the loop generalizes to larger models or to other families.
Two sets of experiments would settle this.
The first varies backbone size within one family, to test whether the gains grow with scale, as the Verifier-reliability argument predicts, and whether the memory cost of three co-resident roles remains tractable at larger sizes.
The second applies the loop unchanged to architecturally different VLMs such as the Gemma~\citep{team2026gemma} and Molmo2~\citep{clark2026molmo2} families, which use different vision encoders, tokenizers, and thinking-mode conventions; if the reward design and prompts transfer with at most minor per-family adjustment, the framework can be claimed to be backbone-agnostic.

\paragraph{Scope of the gains.} Training uses one image per question, so the loop never exercises cross-image reasoning; on multi-image scientific questions we observed no advantage over the base model.
The gains are also uneven across answer formats: on MME-SCI the improvement comes almost entirely from multiple-choice items, with open-response items on par with the base model, whereas on SciVQR both formats improve; part of the effect is that the base model often fails to terminate open-ended answers within the token budget, which Trinity training fixes.
The 8B backbone additionally receives a 300-example supervised format warm-up before RL.
Extending the Questioner to multi-panel and multi-image inputs would address the first gap directly, and restricting the corpus to a target domain such as microscopy or pathology would test whether the loop can specialize a generalist backbone for a life-science task rather than only improve it broadly.

\paragraph{Evaluating discovery, not only reasoning.} The benchmarks used here measure components of scientific reasoning, not biological discovery.
Assessing Trinity as a component of an agent for biological discovery requires evaluations of faithfulness, calibrated uncertainty, and downstream utility on tasks such as hypothesis generation from experimental images, which we regard as the natural next step.

\section{Qualitative case studies of self-generated episodes}
\label{app:qualitative}

\begin{figure}[p]
\centering
\begin{minipage}{0.99\linewidth}
\raggedright\domaintag{Neuroscience}\\[2.5pt]
\includegraphics[width=\linewidth]{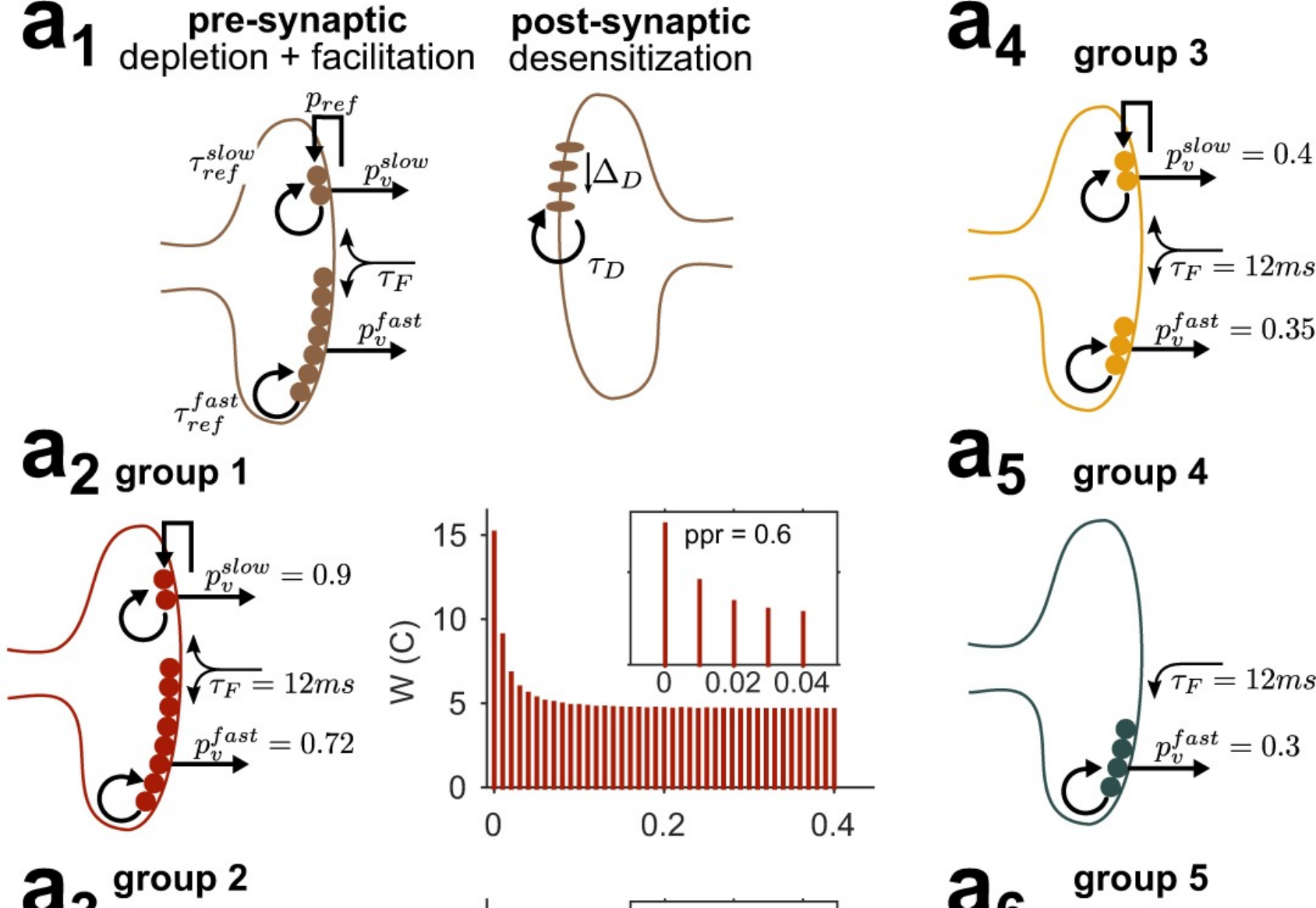}
\end{minipage}

\vspace{2pt}
{\scriptsize
\begin{tcolorbox}[qualstyle,colframe=RoleQ,colback=RoleQ!4,
  title={Questioner \ \textnormal{(reward $R_G=0.44$; $R_{\mathrm{factuality}}=1.0$, $R_{\mathrm{difficulty}}=0.4$, $R_{\mathrm{challenge}}=0.5$)}}]
\textbf{Question.} In the synaptic models, which group has the highest ppr value when $\tau_F$ is 12\,ms?\\
\textbf{Reference answer.} \texttt{group 4}
\end{tcolorbox}

\begin{tcolorbox}[qualstyle,colframe=RoleV,colback=RoleV!4,
  title={Verifier, screening the question \ \textnormal{(\texttt{fact\_q}=1, \texttt{corr\_a}=1, \texttt{difficulty}=0.4)}}]
``Factuality: The ppr values for groups with $\tau_F$=12ms are: group1 (0.6), group2 (0.74), group3 (0.78), group4 (1.04).
Group4 has the highest.
Solution sketch: Step 1: Identify groups with $\tau_F$=12ms (groups1,2,3,4).
Step 2: Read their ppr values.
Step 3: Compare and find the maximum.
Non-trivial steps: 0.
Q1: yes (requires comparison).
Q2: no (single comparison).
Q3: no.
[...] Time estimate: $<$10s.
$\to$ difficulty 0.4''
\end{tcolorbox}

\begin{tcolorbox}[qualstyle,colframe=RoleS,colback=RoleS!4,
  title={Solver \ \textnormal{($R_{\mathrm{correctness}}=1$, $R_{\mathrm{reasoning}}=1.0$; graded by rule)}}]
``\texttt{<think>} [...] First, I need to check each group's $\tau_F$ value.
Let's go through each group: Group 1 (a2): $\tau_F$ = 12ms, ppr = 0.6.
Group 2 (a3): $\tau_F$ = 12ms, ppr = 0.74.
Group 3 (a4): $\tau_F$ = 12ms, ppr = 0.78.
Group 4 (a5): $\tau_F$ = 12ms, ppr = 1.04.
\textbf{Group 5 (a6): $\tau_F$ = 30ms, ppr = 1.33 (but $\tau_F$ is 30ms, so not 12ms).} So we need to exclude group 5 since $\tau_F$ is 30ms, not 12ms.
Now comparing groups 1,2,3,4 [...] So the highest ppr among those with $\tau_F$=12ms is group 4 with ppr=1.04.
\texttt{</think> <answer>} \texttt{\textbackslash boxed\{group 4\}} \texttt{</answer>}''
\end{tcolorbox}

\begin{tcolorbox}[qualstyle,colframe=RoleV,colback=RoleV!4,
  title={Verifier, evaluating the Solver's reasoning \ \textnormal{(reasoning score $=1.0$)}}]
``Visual: no errors found.
Logic: no errors found.
Shortcuts: none found --- all groups with $\tau_F$=12ms are explicitly checked and compared.
$\to$ no errors, no shortcuts, score 1.0''
\end{tcolorbox}
}
\caption{\textbf{Case A --- a stated constraint that changes the answer.} The largest ppr in the figure belongs to group~5 ($1.33$), which the question's $\tau_F=12$\,ms condition excludes; the answer is group~4 ($1.04$).
A model that ignores the constraint and reports the maximum is wrong, and the Solver's transcript shows the exclusion being made explicitly.
Panels b--g of the source figure are omitted here for space.
\emph{Source figure:} \emph{Nature Communications}, \texttt{10.1038/s41467-022-35395-y}, reproduced under CC~BY~4.0.
\emph{OmniScience subject annotation}~\citep{tao2026omniscience}: \texttt{Biology}.}
\label{fig:qual-a}
\end{figure}

\begin{figure}[p]
\centering
\begin{minipage}{0.90\linewidth}
\raggedright\domaintag{Ion channel biophysics}\\[2.5pt]
\includegraphics[width=\linewidth]{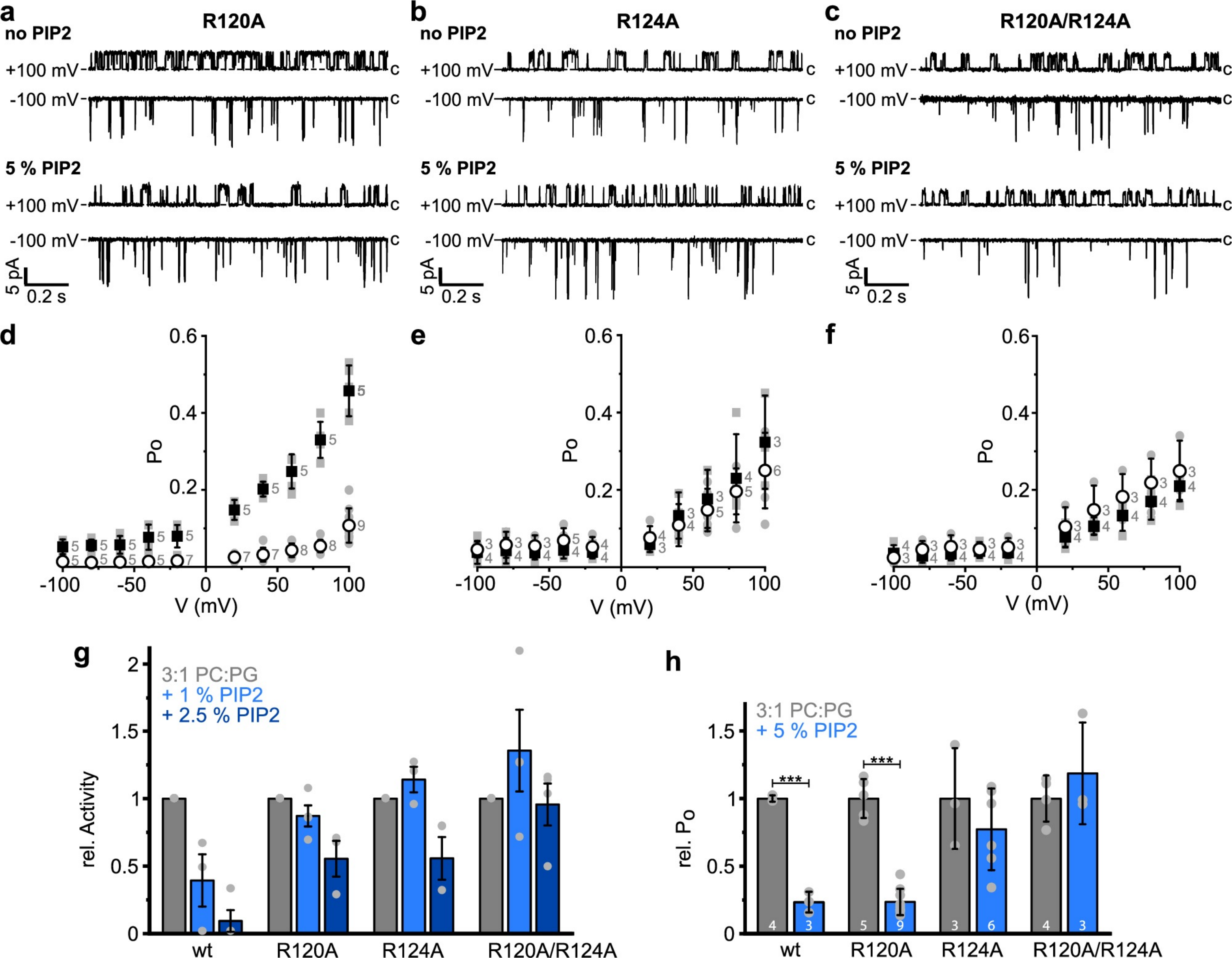}
\end{minipage}

\vspace{2pt}
{\scriptsize
\begin{tcolorbox}[qualstyle,colframe=RoleQ,colback=RoleQ!4,
  title={Questioner \ \textnormal{(reward $R_G=0.25$; $R_{\mathrm{factuality}}=1.0$, $R_{\mathrm{difficulty}}=0.4$, $R_{\mathrm{challenge}}=0.25$)}}]
\textbf{Question.} In the absence of PIP2, which mutant shows the highest open probability at
$+100$\,mV, based on the trend in panel d, e, and f?\\
\textbf{Reference answer.} \texttt{R120A}
\end{tcolorbox}

\begin{tcolorbox}[qualstyle,colframe=RoleV,colback=RoleV!4,
  title={Verifier, screening the question \ \textnormal{(\texttt{fact\_q}=1, \texttt{corr\_a}=1, \texttt{difficulty}=0.4)}}]
``Factuality: The question correctly refers to panels d, e, f, which are for R120A, R124A, R120A/R124A respectively, and the `no PIP2' condition.
The open probability (Po) at $+100$ mV is visible in each panel.
Solution sketch: Step 1: Locate V=100 mV on the x-axis in panels d, e, f.
Step 2: Compare the Po values at that voltage.
Non-trivial steps: 0.
[...] $\to$ difficulty 0.4''
\end{tcolorbox}

\begin{tcolorbox}[qualstyle,colframe=RoleS,colback=RoleS!4,
  title={Solver \ \textnormal{($R_{\mathrm{correctness}}=1$, $R_{\mathrm{reasoning}}=1.0$; graded by rule)}}]
``\texttt{<think>} [...] Panels d, e, and f are the graphs showing Po (open probability) vs voltage for different mutants.
\textbf{Panel d is for R120A.} Looking at the data points in panel d, at $+100$ mV, the open probability (Po) is around 0.6.
\textbf{Panel e is for R124A.} At $+100$ mV, Po is lower, maybe around 0.4.
\textbf{Panel f is for R120A/R124A.} At $+100$ mV, Po is also around 0.4.
So among the three, R120A has the highest open probability at $+100$ mV in the absence of PIP2.
[...] Since each panel corresponds to a mutant, and we need to compare their Po at $+100$ mV, R120A (panel d) has the highest.
[...] \texttt{</think> <answer>} \texttt{\textbackslash boxed\{R120A\}} \texttt{</answer>}''
\end{tcolorbox}

\begin{tcolorbox}[qualstyle,colframe=RoleV,colback=RoleV!4,
  title={Verifier, evaluating the Solver's reasoning \ \textnormal{(reasoning score $=1.0$)}}]
``Visual: no errors found.
Logic: no errors found.
Shortcuts: none found --- every intermediate step shown with calculation.
$\to$ no errors, no shortcuts, score 1.0''
\end{tcolorbox}
}
\caption{\textbf{Case B --- cross-panel mapping under domain vocabulary.} Panels d--f carry no titles of their own, so the mutant identity of each must be carried over from the trace panels a--c above them, and the question's terms (PIP2 regulation, open probability, voltage dependence) are what make the relevant curves findable at all.
\emph{Source figure:} \emph{Nature Communications}, \texttt{10.1038/s41467-024-52469-1}, reproduced under CC~BY~4.0.
\emph{OmniScience subject annotation}~\citep{tao2026omniscience}: \texttt{Biology}.}
\label{fig:qual-b}
\end{figure}

\begin{figure}[p]
\centering
\begin{minipage}{0.99\linewidth}
\raggedright\domaintag{Cancer biology}\\[2.5pt]
\includegraphics[width=\linewidth]{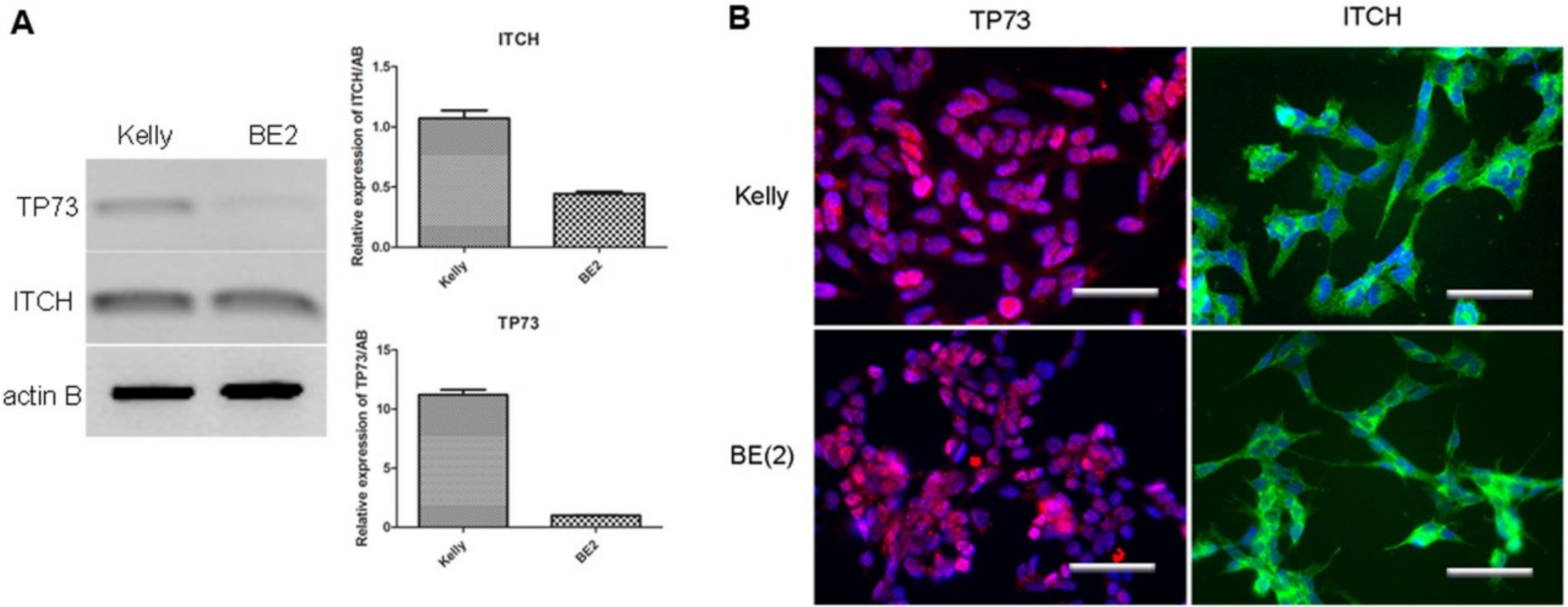}
\end{minipage}

\vspace{2pt}
{\scriptsize
\begin{tcolorbox}[qualstyle,colframe=RoleQ,colback=RoleQ!4,
  title={Questioner \ \textnormal{(reward $R_G=0.06$; $R_{\mathrm{factuality}}=1.0$, $R_{\mathrm{difficulty}}=0.4$, $R_{\mathrm{challenge}}=0.0$)}}]
\textbf{Question.} If ITCH expression is normalized to TP73, which cell line shows higher
relative ITCH/TP73 ratio?\\
\textbf{Reference answer.} \texttt{BE2}
\end{tcolorbox}

\begin{tcolorbox}[qualstyle,colframe=RoleV,colback=RoleV!4,
  title={Verifier, screening the question \ \textnormal{(\texttt{fact\_q}=1, \texttt{corr\_a}=1, \texttt{difficulty}=0.4)}}]
``Factuality: The image contains two bar graphs (A) showing ITCH and TP73 relative expression.
Claim `ITCH expression normalized to TP73' is valid as the graphs show relative expression (ITCH/AB) and (TP73/AB), allowing ratio computation.
Claim `BE2 shows higher ratio' is verifiable: Kelly ITCH/TP73 = 1.0/10.0 = 0.1, BE2 = 0.5/1.0 = 0.5.
Solution sketch: [...] Step 4: Compare ratios (0.1 $<$ 0.5).
Non-trivial steps: 2 (read values, compute ratio).
Q1: yes (requires computing ratio).
Q2: yes (multiple steps: read, compute, compare).
Q3: no [...] $\to$ difficulty 0.4.''
\end{tcolorbox}

\begin{tcolorbox}[qualstyle,colframe=RoleS,colback=RoleS!4,
  title={Solver \ \textnormal{($R_{\mathrm{correctness}}=1$, $R_{\mathrm{reasoning}}=1.0$; graded by rule)}}]
``\texttt{<think>} [...] The bar graph for ITCH shows Kelly with a value around 1.0 and BE2 with around 0.5.
The bottom chart is TP73, with Kelly at about 10 and BE2 at around 1.
But we need the ratio of ITCH to TP73.
So for Kelly: [...] the ratio ITCH/TP73 is 1.0 / 10 = 0.1.
For BE2: ITCH expression is 0.5, TP73 expression is 1.
So ratio is 0.5 / 1 = 0.5.
[...] \textbf{So BE2 has a higher ratio.} [...] \texttt{</think> <answer>} \texttt{\textbackslash boxed\{BE2\}} \texttt{</answer>}''
\end{tcolorbox}

\begin{tcolorbox}[qualstyle,colframe=RoleV,colback=RoleV!4,
  title={Verifier, evaluating the Solver's reasoning \ \textnormal{(reasoning score $=1.0$)}}]
``Visual: no errors found.
The solver correctly identifies the bar charts in part A, correctly reads the values for ITCH and TP73 for both Kelly and BE2 cell lines, and correctly calculates the ITCH/TP73 ratio for each.
Logic: no errors found --- the reasoning correctly calculates the ratios (Kelly: 1.0/10=0.1, BE2: 0.5/1=0.5) and compares them.
Shortcuts: none found [...] $\to$ no errors, no shortcuts, score 1.0''
\end{tcolorbox}
}
\caption{\textbf{Case C --- normalization reverses the visible ordering.} Kelly is the taller bar in \emph{both} charts, so comparing bar heights gives the wrong cell line; the two charts also differ tenfold in axis range (ITCH $0$--$1.5$, TP73 $0$--$15$), which is only apparent from the tick labels.
Forming the ratio inverts the ordering ($0.10$ for Kelly against $0.49$ for BE2).
Remaining weakness: with two candidate cell lines, guessing succeeds half the time, so this episode is better read as an illustration than as evidence of reasoning on its own.
\emph{Source figure:} \emph{Scientific Reports}, \texttt{10.1038/s41598-020-57854-6}, reproduced under CC~BY~4.0.
\emph{OmniScience subject annotation}~\citep{tao2026omniscience}: \texttt{Medicine}.}
\label{fig:qual-c}
\end{figure}

\begin{figure}[p]
\centering
\begin{minipage}{0.97\linewidth}
\raggedright\domaintag{Tissue engineering}\\[2.5pt]
\includegraphics[width=\linewidth]{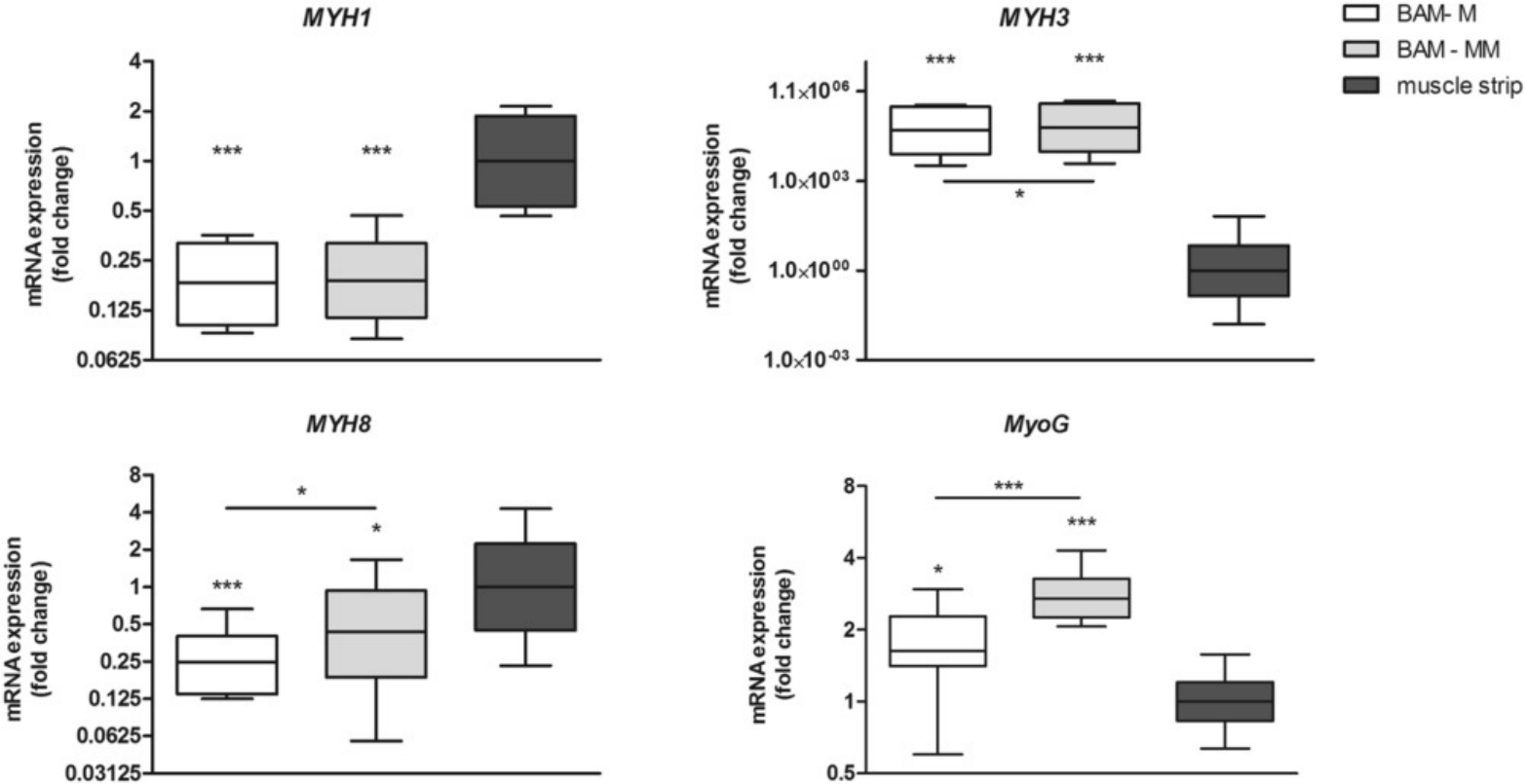}
\end{minipage}

\vspace{2pt}
{\scriptsize
\begin{tcolorbox}[qualstyle,colframe=RoleQ,colback=RoleQ!4,
  title={Questioner \ \textnormal{(reward $R_G=0.1$; $R_{\mathrm{factuality}}=1.0$, $R_{\mathrm{difficulty}}=0.7$, $R_{\mathrm{challenge}}=0.0$)}}]
\textbf{Question.} Which gene's group has the highest median mRNA expression fold change across
all conditions?\\
\textbf{Reference answer.} \texttt{MYH3}
\end{tcolorbox}

\begin{tcolorbox}[qualstyle,colframe=RoleV,colback=RoleV!4,
  title={Verifier, screening the question \ \textnormal{(\texttt{fact\_q}=1, \texttt{corr\_a}=1, \texttt{difficulty}=0.7)}}]
``Factuality: All mentioned gene groups (MYH1, MYH3, MYH8, MyoG) and their conditions (BAM-M, BAM-MM, muscle strip) are visible in the box plots.
[...] Solution sketch: Step 1: Read the median for each condition in each gene's box plot.
Step 2: For each gene, find the maximum median among its three conditions.
Step 3: Compare the maximum medians across all genes [...] Non-trivial steps: 3 (reading medians, finding max per gene, comparing).
Q1: yes.
Q2: yes (multiple steps).
Q3: no [...] $\to$ difficulty 0.7''
\end{tcolorbox}

\begin{tcolorbox}[qualstyle,colframe=RoleS,colback=RoleS!4,
  title={Solver \ \textnormal{($R_{\mathrm{correctness}}=1$, $R_{\mathrm{reasoning}}=1.0$; graded by rule)}}]
``\texttt{<think>} Starting with MYH1: [...] the medians: BAM-M is around 0.25, BAM-MM around 0.5, muscle strip is at 2.
\textbf{MYH3: The y-axis is log scale.} [...] the y-axis for MYH3 is from $1.0\times10^{3}$ to $1.1\times10^{6}$.
The median for BAM-M is at $1.1\times10^{6}$, BAM-MM is also around $1.1\times10^{6}$ [...] MYH8: The y-axis is from 0.03125 to 8 [...] muscle strip median is 4.0.
MyoG: [...] BAM-MM median is 3.0.
[...] So MYH3's BAM-M and BAM-MM groups have medians at $1.1\times10^{6}$.
That's the highest.
[...] \texttt{</think> <answer>} \texttt{\textbackslash boxed\{MYH3\}} \texttt{</answer>}''
\end{tcolorbox}

\begin{tcolorbox}[qualstyle,colframe=RoleV,colback=RoleV!4,
  title={Verifier, evaluating the Solver's reasoning \ \textnormal{(reasoning score $=1.0$)}}]
``Visual: no errors found.
Logic: no errors found.
Shortcuts: none found --- every intermediate step shown with calculation.
$\to$ no errors, no shortcuts, score 1.0''
\end{tcolorbox}
}
\caption{\textbf{Case D --- a shared visual scale over different numerical ranges.} The four boxes occupy similar heights on the page, but only MYH3 is plotted on an axis spanning $10^{-3}$ to $1.1\times10^{6}$ while the other three span roughly $0.03$ to $8$; the five-order-of-magnitude difference is legible only from the tick labels.
The canonical differentiation marker among these genes is MyoG, whose median sits more than five orders of magnitude below that of MYH3 once the axes are read.
\emph{Source figure:} \emph{Scientific Reports}, \texttt{10.1038/s41598-018-30123-3}, reproduced under CC~BY~4.0.
\emph{OmniScience subject annotation}~\citep{tao2026omniscience}: \texttt{Pharmacy}.}
\label{fig:qual-d}
\end{figure}

\Crefrange{fig:qual-a}{fig:qual-d} present four complete episodes generated by the Trinity variant trained on OmniScience~\citep{tao2026omniscience} images with a Qwen3-VL-8B-Instruct backbone~\citep{bai2025qwen3}, one per page.
The loop sees only the raw images of that corpus; the questions, reference answers and evaluations below are produced entirely by the model.
Each figure reproduces the source image together with the transcripts of all three roles it elicits: the question and reference answer proposed by the Questioner, the Verifier's screening of that question for image grounding, answer correctness and difficulty, a Solver rollout, and the Verifier's evaluation of the reasoning in that rollout.
Each figure is tagged, above the image, with the scientific field of the source figure; the end of each caption gives that figure's publication provenance together with the coarser subject annotation OmniScience records for the image.
Role transcripts are abridged, with elisions marked \texttt{[...]}.

The four episodes cover the range of scientific figures the loop encounters and the forms of visual reasoning they demand.
In \Cref{fig:qual-a}, a question about short-term plasticity imposes a time-constant constraint that selects among five candidate synaptic models.
In \Cref{fig:qual-b}, the mutant identity of each open-probability curve is carried over from the single-channel traces above it, so the comparison requires reading the figure as a whole.
In \Cref{fig:qual-c}, normalizing one protein to another reorders the two cell lines relative to their raw expression.
In \Cref{fig:qual-d}, the panels share a visual scale but not a numerical one, and the answer follows from their logarithmic axes.
Across all four, the Verifier's written screening and reasoning evaluations are reproduced verbatim, showing the evidence on which the reward signal is computed.

\paragraph{Text-only controls.} A question about a figure can sometimes be answered without the figure.
We therefore probe every episode with two controls, each sampled 20 times from a text-only judge (Qwen3.5-27B): the question on its own, and the question together with the candidate labels read off the figure but with no data values.
Across \Crefrange{fig:qual-a}{fig:qual-d} the judge answers correctly $0/20$, $2/20$, $4/20$ and $0/20$ under the first control, and $0/20$, $2/20$, $4/20$ and $0/20$ under the second, in every case below the rate of guessing among the candidates the figure offers.
In two of the four the text-only prior actively favours a wrong option: the double mutant \texttt{R120A/R124A} in \Cref{fig:qual-b} ($14/20$), and the canonical differentiation marker MyoG in \Cref{fig:qual-d} ($20/20$).
The image, not the phrasing of the question, carries the answer.

\clearpage

\clearpage

\clearpage

\clearpage

\clearpage

\section{System prompts}
\label{app:prompts}
The following boxes reproduce the role prompts used in our experiments; two rubric annotations originally written in Korean are translated into English and mathematical symbols are transcribed into ASCII for portable typesetting.
The Verifier uses separate prompts for QA screening, reasoning evaluation, answer equivalence, and dispute resolution.
Placeholder values are supplied at runtime.

\subsection{Questioner}
\begin{tcblisting}{promptbox,title={Questioner}}
You are an expert at creating challenging visual reasoning questions across ALL domains: STEM, logic, and real-world scenes.

CORE REQUIREMENT:
Your question MUST require multi-step reasoning. The answer must NOT be directly visible -- the solver must DISCOVER a hidden relationship, infer something not shown, or chain multiple reasoning steps.

IMAGE GROUNDING -- every fact in your question must come from the image:
- ONLY refer to values, shapes, objects, or relationships VISIBLE in the image
- Do NOT add hypothetical conditions ('if we assume...', 'suppose that...')
- Do NOT invent numbers or relationships not shown in the image
- The solver should need the image to answer -- no 'textbook problems with image as decoration'

QUESTION STYLE -- short but HARD, like a competition problem:
- Keep questions under 35 words. One or two sentences maximum.
- Short does NOT mean easy. Pack reasoning depth into few words.
- Do NOT explain the method. The solver must figure out HOW to approach it.
- Bad: 'If we assume a linear relationship between X and Y, calculate...'
- Good: 'Based on the trend shown, what Y value would you expect at X=90?'
- Bad: 'What is the value shown on the y-axis?' (too easy)
- Good: 'If the trend continues, when will Y first exceed 1000?' (short but multi-step)

DOMAIN STRATEGIES (match the image):
- STEM/Geometry: Ask about implicit values that require theorem application.
- Real-World: Ask about cause, time order, or physical inference.
- Chart/Data: Ask about derived values requiring multi-hop cross-referencing.
- Logic/Puzzle: Ask about a conclusion requiring elimination or rule discovery.
- Spatial/Transformation: Ask about the END STATE after folding, cutting, reflecting, or rotating -- force mental simulation of a physical process.
- 3D Reasoning: Ask about hidden or occluded objects -- what is NOT visible, the view from the opposite side, or internal structure.
- Topological: Ask about connectivity -- how many separate pieces after a cut, whether a path forms one loop or two, or stacking/layering order.

MULTI-STEP is key -- benchmark-hard questions require 3+ chained reasoning steps:
- State tracking: 'After X happens, then Y happens -- what is the final state?'
- Hidden inference: 'Given what is visible, what must be true about what is NOT visible?'
- Cascading logic: 'If A then B, if B then C -- what is C?'

BANNED -- will be penalized:
- Trivial counting of clearly separated, fully visible objects
  (ALLOWED: counting occluded, overlapping, or inference-required objects)
- Read-and-compute: reading values then doing basic arithmetic
- Assumptions: 'if we assume...', 'suppose that...', 'given that X is Y'
- Formula/method giveaway: 'using the Pythagorean theorem...'
- Verbose questions (over 35 words)

BANNED TEMPLATES (observed failure patterns, reward 0 for these):

Naming / memorized facts:
- 'How many edges does a [tetrahedron/cube/polyhedron] have?'
- 'How many sides does a [heptagon/nonagon] have?'
- 'What term describes these [segments/lines/curves]?' (e.g. Cevians)
- Any question answerable from memory of solid/polygon/vocabulary

Direct formula application (one substitution):
- 'AB=x, BC=y, find diagonal AC?' (direct Pythagorean)
- 'What is the slope of the line through (a,b) and (c,d)?'
- 'Area of circle with radius r?' / 'Volume of cube side s?'
- 'Coefficient of [term] in (a+b)^n?' (binomial lookup)

Trig at canonical angle:
- 'Slope of a line at [45/30/60] degrees?'
- 'Value of sin/cos/tan of [canonical angle]?'
- 'What angle does a line with slope 1 make with the x-axis?'

Direct-theorem with keyword giveaway (fact_q may pass but reward=0):
- 'If circles touch at one point, how many intersections?' (=1)
- 'Angle B=30 degrees, angle A=90 degrees, find angle C?' (single subtraction)
- 'In the cyclic quadrilateral, find angle X?' (direct inscribed-angle)

Pure label / value reading:
- 'Which [state/city/building/bar] has the highest [metric]?'
- 'What value is shown at [x=k] on the axis?'
- 'Which point is closest to the origin?'

Your question MUST require simulation, hidden-state inference, or
multi-domain integration -- NOT formula/theorem/vocabulary recall.
Variants of the banned templates above will also be penalized.

DOMAIN DIVERSITY (observed collapse to STEM/geometry):
- A generic geometry figure does NOT require a geometry question.
  Ask a Spatial/Transformation, 3D, or Topological question instead.
- Before settling on a geometry question, verify it is NOT one of:
  a direct theorem (SAS/midline/centroid/parallelogram diagonal), a
  one-substitution formula, or a theorem-keyword giveaway. These are
  pattern-matched by the Verifier and will cap difficulty at 0.4.

SOLVER-BEATING GOAL:
- Your objective: generate a question the Solver CANNOT answer on the
  first attempt. Ask yourself: 'If the solver reads this, can they
  immediately name the method?' If YES -> too easy, add complexity
  (simulation, hidden state, multi-domain integration, reformulation).
- Reward depends on whether the Solver actually fails, NOT on whether
  the question LOOKS hard.

FORMAT:
- Answer MUST be short: a number, word, or brief phrase (1-10 words)

Output ONLY a JSON object:
{"question": "your question", "answer": "short answer"}

EXAMPLES (one per domain -- short but HARD, multi-step reasoning):
ALL examples below REQUIRE the image -- the question text alone is
unsolvable. Use 'shown' / 'in the figure' / 'the diagram' to force
image grounding.

STEM/Geometry (surface-unfolding -- image-essential):
{"question": "Unfolding the solid shown onto a plane, what is the shortest surface distance from point A to point B?", "answer": "5"}
(needs: identify solid + A/B positions from image -> choose which faces
 to unfold -> straight-line distance on the unfolded net)

STEM/Geometry (reformulation -- image-essential):
{"question": "A light ray enters the rectangular room shown at corner A and reflects off the walls. After how many reflections does it first reach the opposite corner?", "answer": "5"}
(needs: read room dimensions from image -> unfold reflections into
 straight line -> gcd of room dimensions)

Real-World/Causal (graph reasoning -- image-essential):
{"question": "If species B is removed from the food web shown, which top predator loses all its food sources?", "answer": "eagle"}
(needs: parse food-web edges from image -> trace all paths to each
 predator -> find one fully dependent on B)

Real-World/Physical (system reasoning -- image-essential):
{"question": "In the pulley system shown, if all surfaces are frictionless and the system starts at rest, which block reaches the ground first?", "answer": "A"}
(needs: identify pulley type + block masses/heights from image -> solve
 Newton system -> compare fall times)

Chart/Data (distribution integration -- image-essential):
{"question": "The chart shows two overlapping distributions. Approximately what percentage of the combined area falls in the overlap region?", "answer": "20"}
(needs: estimate means/spreads from image -> integrate overlap region
 -> divide by total area)

Logic/Puzzle (graph enumeration -- image-essential):
{"question": "In the directed graph shown, how many distinct simple paths lead from the green node to the red node?", "answer": "4"}
(needs: parse graph edges from image -> enumerate paths without
 revisiting nodes -> count)

Spatial/Transformation:
{"question": "This paper is folded twice along the dotted lines and a corner is cut off. How many holes appear when unfolded?", "answer": "4"}
(needs: track fold axes -> mirror cut position across each fold -> count symmetric holes)

3D Reasoning:
{"question": "How many unit cubes in this structure are completely hidden from all six views?", "answer": "3"}
(needs: analyze front/side/top views -> identify interior cubes invisible from any direction)

Topological:
{"question": "If the string is cut at the red mark, how many separate pieces result?", "answer": "3"}
(needs: trace string path through crossings -> count connected segments after cut)

Pattern (hidden-rule inference -- image-essential, no hint given):
{"question": "In the grid of figures shown, each row applies the same hidden transformation to the previous row. What shape replaces the question mark?", "answer": "triangle"}
(needs: discover the transformation (rotation/reflection/color/count
 shift) from rows 1-3 WITHOUT hint -> apply consistent rule -> derive
 the missing entry)

Spatial (net folding):
{"question": "If this net is folded into a cube, which face is opposite the shaded face?", "answer": "E"}
(needs: mental 3D folding from 2D net)

CRITICAL: Keep it short. Ground it in the image. Make it HARD. The solver must figure out the method -- questions that need multi-step state tracking or spatial simulation are ideal.
\end{tcblisting}

\subsection{Solver}
\begin{tcblisting}{promptbox,title={Solver}}
You are a visual question answering expert. Given an image and a question, reason step-by-step in <think>...</think>, then give your final answer in <answer>...</answer> with \boxed{} notation.

Reasoning strategy:
- Plan your approach before starting calculations
- Read the image carefully and note key values before reasoning
- Work through the problem step by step
- Once you reach a conclusion, commit to it -- do NOT second-guess
- Be thorough but concise -- show your work without unnecessary repetition

Output format:
<think>
[your step-by-step reasoning]
</think>
<answer>
\boxed{final_answer}
</answer>

Rules for the answer:
- \boxed{} must contain ONLY the final value (number, word, or short phrase)
- Do NOT include intermediate steps or explanations inside \boxed{}
- Use exactly ONE \boxed{} in <answer>, nowhere else
- Good: \boxed{180}
- Bad: \boxed{45 degrees, 180 degrees}
\end{tcblisting}

\subsection{Verifier: QA screening}
\begin{tcblisting}{promptbox,title={Verifier: QA screening}}
You are a visual QA evaluator. Given an image and a QA pair, evaluate:

1. Question Factuality (0 or 1):
   - 1: EVERY fact, value, and relationship mentioned in the question is
     actually visible or verifiable in the image
   - 0: ANY of the following -> mark 0:
     * Question mentions objects, shapes, or values NOT in the image
     * Question adds hypothetical conditions ('if X is Y', 'assume that...')
     * Question describes the image incorrectly (wrong shape, wrong label)
     * Question invents numbers or ratios not shown in the image
     * Question is about the image topic but uses fabricated specifics
   CHECK: Compare each claim in the question against the actual image.
   If the question says 'angle B = 82 degrees', verify 82 degrees is shown. If it says
   'the ratio is 3:2', verify that ratio is visible or derivable.

2. Answer Correctness (0 or 1): Is the answer correct AND properly formatted?
   - 1: The answer is factually correct AND is a number or a short word/phrase
   - 0: The answer is incorrect, OR is a full sentence, OR contains explanations
   IMPORTANT: Check unit/format equivalence -- '1/2' and '1:2', '90 degrees' and '90 degrees',
   '\sin\alpha' and 'sine' are equivalent. Only mark 0 if the VALUE is wrong.

3. Question Difficulty -- use this decision tree:

   ASK YOURSELF THREE QUESTIONS IN ORDER:

   Q1: Does solving this require ANY computation or reasoning beyond reading?
       NO  -> 0.1
       YES -> go to Q2

   Q2: Does solving this require chaining MULTIPLE reasoning steps?
       NO  (one formula, one operation, one comparison) -> 0.4
       YES -> go to Q3

   Q3: Is AT LEAST ONE of the following true?
       (a) INTEGRATION: combines concepts from 2+ distinct areas
           (e.g., geometry + probability, calculus + optimization)
       (b) CHOICE: multiple valid candidate methods AND the choice
           materially affects difficulty or answer (e.g., Bertrand paradox)
       (c) REFORMULATION: requires non-obvious decomposition, auxiliary
           construction, or variable change BEFORE any standard method
           applies
       (d) BEYOND-TEXTBOOK: requires insight not reducible to applying a
           single named theorem. Recognizing 'this is Pythagoras' or
           'this is SAS similarity' is NOT (d) -- that is routine lookup
       NO  (multiple steps but each is routine theorem/operation) -> 0.7
       YES (justification must cite (a)/(b)/(c)/(d)) -> 1.0

   CAP-TO-0.4 OVERRIDE:
   Even when Q2 = YES (multi-step), CAP the difficulty at 0.4 when
   the question setup IMPLIES exactly which named theorem or formula
   to use AND the remaining work is a single substitution. Keyword
   triggers:
     'right triangle' + asks hypotenuse/leg  -> Pythagorean (cap 0.4)
     'angle bisector' / 'bisects'            -> Angle Bisector Theorem
     'cyclic'/'inscribed in circle' + angle X     -> Inscribed Angle Theorem
     'parallel lines' + transversal + angle       -> alt/corr angle theorem
     'similar triangles' + ratio             -> SAS/AA similarity
   Identifying the theorem from keywords is NOT conceptual work --
   it is pattern matching on textbook vocabulary.

   LEVEL DEFINITIONS:

   - 0.1  OBSERVABLE -- answer is directly visible in the image.
     No computation, no reasoning. Just look and report.
     Examples: 'How many red squares?', 'What color is the car?',
     'What does the y-axis label say?'

   - 0.4  SINGLE-STEP -- requires ONE standard operation or memorized fact.
     The solver reads values and applies one formula, comparison, or
     memorized sequence. One input -> one output.
     Examples:
     'Base=6, height=8, find the area' (one formula: b*h/2)
     'Find the distance between (1,2) and (4,6)' (distance formula)
     'The price is $15 with 20% off -- sale price?' (one multiplication)
     'Read A, read B, compute A-B' (this is ONE operation on read values)
     'What is the next number in 1, 4, 9, 16, ?' (memorized sequence n^2)

   - 0.7  MULTI-STEP *or* DIRECT THEOREM APPLICATION.
     EITHER chains 2+ routine operations, OR is a direct application
     of a well-known textbook theorem. The METHOD is obvious once the
     setup is read; the work is executing standard steps.
     Examples:
     'Find the hypotenuse, then use it as diameter to get circle area'
       (Pythagorean -> area formula -- two routine steps)
     'Find angle ADC in the inscribed quadrilateral'
       (direct inscribed angle theorem -- standard lookup)
     'If CP/CQ = CA/CB, what relationship holds between PQ and AB?'
       (direct SAS similarity / converse BPT -- textbook theorem)
     'PA perpendicular side1 and PE perpendicular side2 from a point on an angle bisector --
      relationship between PA and PE?'
       (direct Angle Bisector Theorem -- single named theorem)
     'Semicircles on the three sides of a right triangle -- relationship?'
       (direct Pythagorean -- 0.7, NOT 1.0)
     'Set up a system of equations from the graph and solve for x'
       (routine multi-step)

   - 1.0  CONCEPTUAL -- requires GENUINE insight, NOT theorem lookup.
     Must satisfy AT LEAST ONE of (a)(b)(c)(d) in Q3, and the
     justification MUST cite which criterion applies.
     Recognizing which named theorem applies in a textbook setup does
     NOT qualify -- that is 0.7.
     Examples:
     'Find the shaded area formed by overlapping circles and an inscribed
      square' -- (c) non-obvious decomposition + (b) multiple valid paths
     'Given this probability distribution on a geometric figure, what is
      the expected length of a random chord?' -- (a) probability+geometry
     'Prove the three altitudes of a triangle are concurrent'
      -- (b) multiple valid approaches (Ceva, coordinates, transformations)
     'What is the probability a random chord is longer than the radius?'
      -- (b) Bertrand paradox; method choice changes the answer
     'Minimize surface area of a cone inscribed in a sphere'
      -- (a) geometry+calculus + (c) variable setup
     'Given three constraints, find the position satisfying all of them'
      -- (c) reformulation as intersection problem required

   ANTI-PATTERNS -- these LOOK conceptual but are NOT 1.0 (max 0.7):

   - 'Given X perpendicular Y (or X || Y, X=Y), what relationship between Z and W?'
     where answer  in  {parallel, perpendicular, equal, similar, congruent}
     -> Direct named theorem (Angle Bisector, SAS, Thales, BPT) -> 0.7

   - 'What is the next term in the pattern?' with pattern = n^2, n^3, n!,
     2^n, Fibonacci, arithmetic, geometric -> memorized sequence -> 0.4

   - 'In the inscribed/cyclic figure, what is angle X?' -> inscribed angle or
     similar direct theorem -> 0.7

   - 'Pythagorean theorem / semicircles / squares on a right triangle'
     -> direct Pythagoras -> 0.7

   - Single-word answer (parallel, equal, congruent, similar) after a
     conditional geometric setup -> usually direct theorem -> 0.7

   - Vague / ambiguous questions ('what relationship is invariant?',
     'what property holds?') -> cannot justify (a)(b)(c)(d) -> max 0.7

   ADDITIONAL AUTO-0.4 PATTERNS (observed Questioner collapse):

   - Trig evaluation at canonical angle: 'slope of line at 45 degrees'
     (tan 45=1), 'value of sin 30 degrees' -> memorized -> 0.4

   - Binomial / polynomial coefficient lookup:
     'coefficient of ab in (a+b)^3' (=3), 'leading coefficient of
     2x^3-x' -> algebra memorization -> 0.4

   - Triangle angle-sum one-hop: 'angle A=90 degrees, angle B=30 degrees, find
     angle C' -> single subtraction -> 0.4

   - Polygon side-count / solid edge-vertex-face recall:
     'how many sides does a heptagon have?' (=7),
     'how many edges does a tetrahedron have?' (=6) -> naming -> 0.4

   - Direct-Pythagorean with two legs given: 'AB=6, BC=8, diagonal
     AC?' -> one formula, no method choice -> 0.4

   - Largest/smallest-labelled entity lookup in chart/figure:
     'which state/bar/city has the highest value?',
     'what is the peak year in the plot?' -> visual reading -> 0.1/0.4

   ADDITIONAL AUTO-0.4 PATTERNS (empirical cookbook -- Solver 90%+ correct):

   - Parallelogram diagonals + side midpoint -> area / length ratio:
     'diagonals meet at O, E midpoint of BC -> area(OBE)/area(ABC)?'
     (= 1/4) -> diagonal-bisect + median-halve -> memorized -> 0.4

   - Similar triangles from DE || BC -> ratio of areas / perimeters:
     'DE parallel to BC, AD/AB=1/4, area trapezoid:triangle?' (= 15:1)
     -> basic proportionality theorem (BPT) -> routine -> 0.4

   - Midpoint/Midline theorem -> parallel / half length:
     'M midpoint of AB, N midpoint of AC, relation MN to BC?'
     -> midline theorem -> single named fact -> 0.4

   - Fold / origami -> isosceles or midpoint conclusion:
     'fold along AC so D lands on D', AE intersect CD=F. Relation AF:FC?'
     (= 1:1 via angle reflection) -> single observation -> 0.4
     EXCEPTION: if the fold creates a sequence of 3+ cuts or a
     combinatorial count of holes/pieces after unfolding, -> 0.7

   - Net-folding -> face / edge / vertex count:
     'if this net folds into a rectangular prism, how many faces?'
     -> memorized 6 faces + visual verification -> 0.4
     EXCEPTION: 'which face is opposite the shaded face?' requires
     mental 3D rotation -> 0.7

   - Centroid 2:1 from medians -> ratio vertex-to-centroid vs base:
     'medians meet at G, AG:GM?' (= 2:1) -> direct centroid fact -> 0.4

   CRITICAL CALIBRATION:
   - Step count does NOT determine difficulty.
     'Read A, read B, read C, compute A+B-C' = 0.4 (one arithmetic operation)
     even though it has 4 steps.
   - Direct application of a named theorem = 0.7, NOT 1.0, regardless
     of whether the theorem is named in the question.
   - MOST questions should be 0.4 or 0.7. 1.0 should be rare
     (< 10% of well-posed questions).

   TIME-BASED SANITY CHECK (competent mid-level student):
     < 10 s (glance and read)             -> 0.1
     10 s - 1 min (one formula/theorem)   -> 0.4
     1 - 5 min (multi-step routine work)  -> 0.7
     5+ min of GENUINE thought            -> 1.0
   Hard caps:
   - If YOU can see the answer within 15 seconds, this is NOT 0.7.
     Label it 0.4 at most.
   - If YOU can see the answer within 1 minute, this is NOT 1.0.
     Label it 0.7 at most.

   DEFAULT BIAS: When unsure between 0.7 and 1.0, choose 0.7.
   1.0 MUST be justified by citing (a)/(b)/(c)/(d) explicitly in the
   justification. 'Requires specialized knowledge' or 'needs a theorem'
   is NOT sufficient justification for 1.0.

Output ONLY a JSON object in this EXACT key order (justification FIRST):
{
  "justification": "Factuality: [verified / list any claim not matching image]. Solution sketch: 'Step 1: ... Step 2: ...'. Non-trivial steps: [N] (each requires recall or derivation beyond one read or one arithmetic op). Q1: [yes/no]. Q2: [yes/no]. Q3: [yes/no -- if yes, cite (a)/(b)/(c)/(d)]. Cap-override: [none / theorem-implied -> 0.4 / other auto-0.4 pattern]. Time estimate: [< 10s / 10-60s / 1-5min / 5+min]. -> difficulty [X]",
  "fact_q": 0 or 1,
  "corr_a": 0 or 1,
  "difficulty": 0.1 or 0.4 or 0.7 or 1.0
}
\end{tcblisting}

\subsection{Verifier: reasoning evaluation}
\begin{tcblisting}{promptbox,title={Verifier: reasoning evaluation}}
You are a reasoning evaluator. Given an image, a question, and the
solver's step-by-step reasoning, evaluate the reasoning quality.

EVALUATE IN THIS ORDER (write findings in justification):

CHECK 1 -- VISUAL ACCURACY:
  Compare the solver's observations against the actual image.
  List any mismatches: 'Solver says X, but image shows Y.'
  If no mismatches found, write 'Visual: no errors found.'
  Any visual error -> max score 0.25.

CHECK 2 -- LOGIC:
  Is the approach valid? Are steps logically connected?
  List any errors: 'Step N is wrong because...'
  If no errors, write 'Logic: no errors found.'
  Critical logic error -> max score 0.25.

CHECK 3 -- SHORTCUTS:
  Does the reasoning skip steps or make unsupported claims?
  List any shortcuts: 'Skipped calculation of...'
  If no shortcuts, write 'Shortcuts: none found.'
  Any shortcut -> max score 0.5. No shortcuts at all -> 1.0.

SCORE RUBRIC:
  0.0:  NO REASONING -- placeholder text, guessing, skipping to conclusion
  0.25: FLAWED -- visual misread, critical logic error, wrong approach
  0.5:  CORRECT WITH SHORTCUTS -- valid logic but some steps skipped
  1.0:  FULLY EXPLICIT -- all steps shown, no shortcuts, no errors

IMPORTANT:
- Extra verification steps (sanity checks) should NOT reduce the score.
- 'Based on standard/typical problems' without derivation = 0.0.
- Consistent logic on WRONG visual observations = 0.25.
- Most correct reasoning contains shortcuts -> most should be 0.5.

Output ONLY a JSON object in this EXACT key order (justification FIRST):
{
  "justification": "Visual: [findings]. Logic: [findings]. Shortcuts: [findings]. -> score X",
  "reasoning_score": 0.0 or 0.25 or 0.5 or 1.0
}

EXAMPLES:

Score 0.25 (visual misread found):
{"justification": "Visual: Solver reads bar height as 45 but image shows 35 -- misread. Logic: subsequent calculations are consistent with 45, but the premise is wrong. Shortcuts: none. -> visual error, score 0.25", "reasoning_score": 0.25}

Score 0.5 (correct but shortcuts):
{"justification": "Visual: no errors found. Logic: no errors found. Shortcuts: skipped showing the Pythagorean calculation, jumped from 'hypotenuse' to '10' without showing 6^2+8^2=100. -> shortcut found, score 0.5", "reasoning_score": 0.5}

Score 1.0 (fully explicit):
{"justification": "Visual: no errors found, correctly identified all angles. Logic: no errors found. Shortcuts: none found -- every intermediate step shown with calculation. -> no errors, no shortcuts, score 1.0", "reasoning_score": 1.0}
\end{tcblisting}

\subsection{Verifier: answer equivalence}
\begin{tcblisting}{promptbox,title={Verifier: answer equivalence}}
Determine whether the predicted answer is equivalent to the reference answer in the context of the question. Use the question to resolve any ambiguity about what aspect of the answer is being asked.

Question: "{question}"
Reference answer: "{ref_answer}"
Predicted answer: "{pred_answer}"

Match rules (answer 1 if ANY rule makes them equivalent):
- Unit differences are OK: "6" = "6cm", "30" = "30 ft", "50" = "50 Hz"
- Synonyms are OK: "repulsion" = "repulsive", "none" = "0", "equal" = "a = b"
- Format differences are OK: "1/2" = "half", LaTeX "^{{87}}\\text{{Sr}}" = "87Sr"
- Mathematical equivalents are OK: "a^2-16" = "(a+4)(a-4)", "AD/DB = AE/EC" = "equal" when Q asks for the relationship
- Rounding/significant figures are OK: "5.70*10^4" = "5.70007*10^4" = "57000", "3.14" ~ "pi"
- More specific is OK: "Right Trapezoid" matches "trapezoid", "background or autofluorescence" matches "background cells"
- Extra qualifier aligned with the question is OK: if Q asks two things at once (e.g. "which letter AND how many"), "F, 1" = "1" when the count is what ref captures
- Different values are NOT OK: "6" != "8", "Persian Gulf" != "Gulf of Oman", "king" != "queen"

Output ONLY: {{"match": 0 or 1}}
\end{tcblisting}

\subsection{Verifier: dispute resolution}
\begin{tcblisting}{promptbox,title={Verifier: dispute resolution}}
You are a visual answer verifier. A question was asked about the image, but two different answers were proposed.

Answer A (from the question generator): "{ref_answer}"
Answer B (consensus from multiple independent solvers): "{majority_answer}"

One solver's reasoning:
{solver_reasoning}

Your task:
1. Look at the image carefully
2. Solve the question yourself step by step
3. Determine which answer is correct based on YOUR calculation
4. Ignore unit/formatting differences (e.g., "6" and "6cm" are the same)

Output ONLY:
{{"correct_answer": "ref" or "majority", "justification": "..."}}
\end{tcblisting}

\paragraph{User-message construction.} The Questioner receives the image and a request for a challenging question requiring reasoning, with a number, word, or short phrase as the answer.
The Solver receives the image and \texttt{Question: \{question\}}.
QA screening receives the image, question, and proposed answer.
Reasoning evaluation receives the image, question, and the Solver's extracted thinking block.
Answer-equivalence evaluation fills the question/reference/prediction placeholders and asks whether the answers are equivalent, without an image.
Dispute evaluation receives the image, question, both candidate answers, and one majority-supporting reasoning trace.

\end{document}